\documentclass[journal]{IEEEtran}

\usepackage{graphicx}
\usepackage[section]{algorithm}
\usepackage[noend]{algorithmic}
\usepackage{amsthm}
\usepackage{amsmath,amssymb,amsfonts}
\usepackage{bm}
\usepackage{color}
\usepackage{ragged2e}
\usepackage{subfigure}
\usepackage[numbers,sort&compress]{natbib}
\usepackage{mathrsfs}
\usepackage{enumitem}

\usepackage{array}
\newcommand{\PreserveBackslash}[1]{\let\temp=\\#1\let\\=\temp}
\newcolumntype{C}[1]{>{\PreserveBackslash\centering}p{#1}}
\newcolumntype{R}[1]{>{\PreserveBackslash\raggedleft}p{#1}}
\newcolumntype{L}[1]{>{\PreserveBackslash\raggedright}p{#1}}

\renewcommand{\raggedright}{\leftskip=0pt \rightskip=0pt plus 0cm}

\begin{document}

\title{Distributed Deep Learning Model for Intelligent Video Surveillance Systems with Edge Computing}

\author{Jianguo~Chen,
        Kenli~Li,~\IEEEmembership{Senior Member, IEEE},
        Qingying~Deng,\\
        Keqin~Li,~\IEEEmembership{Fellow, IEEE},
        and Philip S. Yu,~\IEEEmembership{Fellow, IEEE}
\IEEEcompsocitemizethanks{
\IEEEcompsocthanksitem Jianguo~Chen, Kenli~Li, and~Keqin~Li are with the College of Computer Science and Electronic Engineering, Hunan University, and the National Supercomputing Center in Changsha, Hunan, Changsha 410082, China.
\protect\\
Corresponding authors: Kenli Li (lkl@hnu.edu.cn) and Keqin Li (lik@newpaltz.edu).
\IEEEcompsocthanksitem Qingying Deng is with the School of Mathematics and Computational Science, Xiangtan University, Xiangtan 411105, China.
\IEEEcompsocthanksitem Keqin Li is also with the Department of Computer Science, State University of New York, New Paltz, NY 12561, USA.
\IEEEcompsocthanksitem Philip S. Yu is with the Department of Computer Science, University of Illinois at Chicago, Chicago, IL 60607, USA, and Institute for Data Science, Tsinghua University, Beijing 100084, China.}
}

\maketitle

\begin{abstract}
In this paper, we propose a Distributed Intelligent Video Surveillance (DIVS) system using Deep Learning (DL) algorithms and deploy it in an edge computing environment.
We establish a multi-layer edge computing architecture and a distributed DL training model for the DIVS system.
The DIVS system can migrate computing workloads from the network center to network edges to reduce huge network communication overhead and provide low-latency and accurate video analysis solutions.
We implement the proposed DIVS system and address the problems of parallel training, model synchronization, and workload balancing.
Task-level parallel and model-level parallel training methods are proposed to further accelerate the video analysis process.
In addition, we propose a model parameter updating method to achieve model synchronization of the global DL model in a distributed EC environment.
Moreover, a dynamic data migration approach is proposed to address the imbalance of workload and computational power of edge nodes.
Experimental results showed that the EC architecture can provide elastic and scalable computing power, and the proposed DIVS system can efficiently handle video surveillance and analysis tasks.
\end{abstract}

\IEEEpeerreviewmaketitle

\section{Introduction}
\IEEEPARstart{V}{ideo} Surveillance (VS) technology has become a fundamental tool for the public and private sector security, such as traffic monitoring, indoor monitoring, and crime and violence detection \cite{tii05, ai06, vs04}.
Edge Artificial Intelligence (EAI) is a promising technology that combines Artificial Intelligence (AI), Internet of Things (IoT), and Edge Computing (EC) technologies \cite{ai01, ai02, ai03}.
Applying EAI technology in VS is an innovative and promising work that migrates computing workloads from the network center to the edge of the network to reduce huge network communication overhead and provide real-time and accurate video analytics solutions.
However, this work is also facing many serious challenges:
(1) how to address the problems of synchronization of distributed AI models in an EC environment;
(2) how to design a feasible edge computing architecture for the VS system, taking into account large-scale monitor terminals, huge video stream, and huge network communication overhead;
and (3) how to keep workload balance among edge nodes under the complicated scenarios of unbalanced connection of monitor terminals and unbalanced computing capacities of edge nodes.

AI technology is widely used in various fields of the intelligent industry, such as intelligent transportation \cite{ai06, tii02}, Internet of Things \cite{tii03, ai08, ec01}, smart grids \cite{tii04, vs01}, and video surveillance \cite{ai02}.
In the field of VS, existing AI and deep learning algorithms, such as Convolutional Neural Network (CNN) and Deep Neural Network (DNN), are mainly used for static image analysis, rather than image streaming and video analysis \cite{ai10, ai11}.
Focusing on distributed VS systems and AI algorithms, most current VS systems rely on traditional centralized or cloud-based solutions, facing huge data communication overhead, high latency, and severe packet loss limitations \cite{vs01, vs04}.
Existing studies have proposed various distributed AI and Deep Learning (DL) algorithms in distributed computing clusters and cloud computing platforms, such as distributed CNN, DNN, and LSTM \cite{ai07, ai08, ai09}.
There are many exploration spaces for distributed AI algorithms and VS system in EC environments \cite{ai01, ai08, vs02}.

In this paper, we focus on intelligent video surveillance systems based on AI and EC technologies, propose a Distributed Intelligent Video Surveillance (DIVS) system using a distributed DL model, and deploy the DIVS system in an edge computing environment.
The contributions of this paper are summarized as follows:

\begin{itemize}
\item We establish a multi-layer edge computing architecture and a distributed DL training model for the DIVS system.
      It migrates computing workloads from the network center to network edges to reduce communication overhead and provide low-latency analysis solutions.
\item We provide task-level and model-level parallel training methods for the distributed DL model.
  In the task-level parallel, multiple DL sub-models with different structures are deployed on each edge node and different data analysis tasks are performed in parallel.
In the model-level parallel, training processes of the CNN model are further parallelized on each edge node.
\item A model parameter updating method is proposed to realize the model synchronization of the global DL model on the EC platform with low communication cost.
\item Considering the unbalanced connection of monitor terminals and unbalanced computing capacities of edge nodes, we propose a dynamic data migration approach to improve the workload balance of the DIVS system.
\end{itemize}

The remainder of this paper is structured as follows.
Section 2 reviews the related work.
Section 3 establishes a multi-layer edge computing architecture and a distributed DL training model for the DIVS system.
The implementation of the proposed DIVS system is described in Section 4.
Experimental evaluation of the DIVS system is presented in Section 5.
Finally, Section 6 concludes the paper.

\section{Related Work}
Various distributed AI and DL algorithms were proposed in distributed computing, cloud computing, fog computing, and edge computing environments to improve their performance and scalability \cite{ai07,ai09, vs02, tii01, ec03, ec02}.
In our previous work, we proposed  a two-layer parallel CNN training architecture in a distributed computing cluster \cite{ai07}.
Li \emph{et al}. discussed the application of Machine Learning (ML) in smart industry and introduced an efficient manufacture inspection system using fog computing \cite{tii01}.
Diro \emph{et al}. proposed a Long Short-Term Memory (LSTM) network for distributed attack detection in fog computing environments \cite{ec03}.
Focusing on edge computing, Khelifi \emph{et al}. discussed the applicability of merging DL models in EC environments, such as CNN, RNN, and RL \cite{ec02}.
In \cite{ec01}, Li \emph{et al}. designed an offloading strategy to optimize the performance of IoT deep learning applications in EC environments.

Focusing on the applications of AI and DL methods in the field of video surveillance, interesting work was presented in \cite{ai02, ai03, ai05, ai06}.
In \cite{ai02}, Ding \emph{et al}. proposed a trunk-branch ensemble CNN platform for video-based face recognition, which can extract complementary information from holistic face images.
In the VS of the public safety field, a temporally memory similarity learning neural network was presented for person re-identification \cite{ai05}.
In the field of traffic monitoring, Zhang \emph{et al}. proposed a CNN-based vehicles detection and annotation algorithm that can identify vehicle positions and extract vehicle properties from video streams \cite{ai06}.
However, most existing methods train the DL models based on static images, and are rarely used for video analysis.

To efficiently handle large-scale video datasets and improve the performance of VS systems, researchers have attempted to deploy VS systems in distributed computing, cloud computing, and edge computing environments \cite{vs01, vs02, tii06}.
In \cite{vs01}, Kavalionak \emph{et al}. introduced a distributed protocol for a face recognition system, which exploits the computing power of the monitoring devices to perform person recognition.
Yi \emph{et al}. built a video analytics system on an EC platform that offloads computing tasks between monitoring devices and edge nodes and provides low-latency video analysis \cite{vs02}.
In \cite{vs03}, Park \emph{et al}. proposed a scalable architecture for an automatic surveillance system using edge computing to reduce cloud resource consumptions and wireless network limitations.

\section{Proposed DIVS System Architecture}

\subsection{Multi-layer EC structure of DIVS System}
\begin{figure}[!ht]
\setlength{\abovecaptionskip}{0pt}
\setlength{\belowcaptionskip}{0pt}
  \centering
  \includegraphics[width=3.4in]{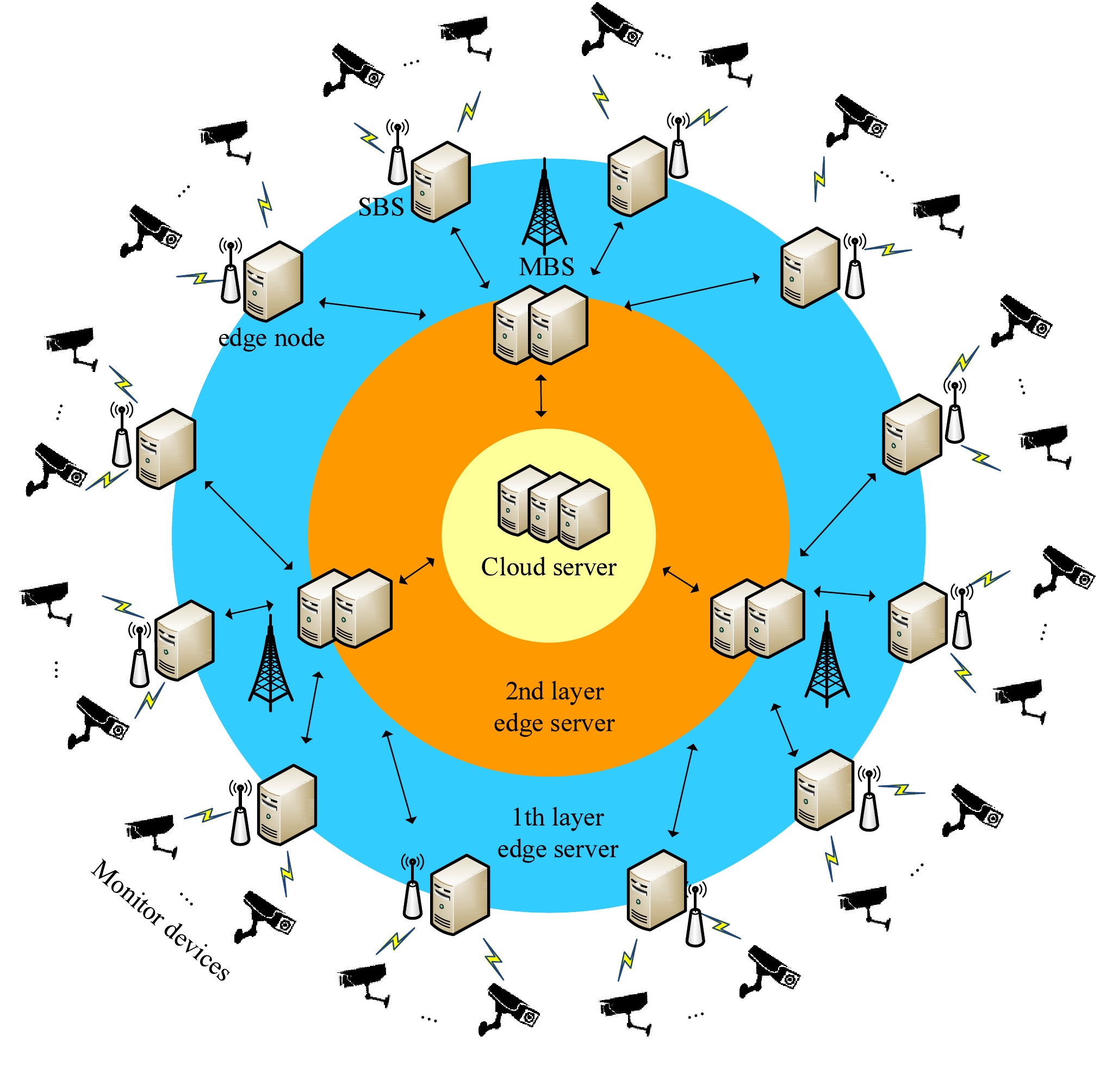}
  \caption{The architecture of the proposed DIVS system.}
  \label{fig01}
\end{figure}

In the IoT and big data era, VS systems hold the characteristics of massive monitoring terminals, wide scope of monitoring, and endless video streams.
At the same time, VS systems face increasing demands of accurate data analysis and low-latency response.
We propose a distributed intelligent video surveillance system by combining the IoT, AI, and EC technologies.
We establish a multi-layer edge computing platform for the DIVS system, providing flexible and scalable computing capabilities and effectively reducing network communication overhead.
The DIVS system consists of a large number of high-definition monitoring devices, multi-layer edge nodes, a cloud server, and a distributed DL model.
The proposed DIVS system architecture is shown in Fig. \ref{fig01}.
The main components of the DIVS system are described as follows.

(1) Monitoring terminals.
There are $N$ Monitoring Terminals (MT, also called monitoring devices) $MTs = \{MT_{1}, ..., MT_{N }\}$ deployed at various positions of monitoring spots, such as traffic road, railway station, airport, or park.
Each MT is equipped with a high-definition camera with a physical resolution of 1080P or 4K and adopts H.264 or H2.65 video coding standards.
Each MT periodically submits the collected video dataset to the edge node to which it belongs by wired or wireless communication.

(2) Edge nodes and cloud server.
Multi-layer Edge Nodes (ENs) are deployed in the DIVS system, which are deployed at different management levels, such as streets, districts, and counties of a city.
The first-level ENs are responsible for connecting the MTs of each monitoring spot.
Each middle-level EN is connected to all low-level ENs within its jurisdiction and is connected the high-level EN to which it belongs.
High-level ENs connect to the cloud server, which provides task scheduling, data management, and resource allocation.

(3) Distributed DL model for video analytics applications.
The video management and analytics application is a core component of the DIVS system, which provides management functions and video analysis functions.
In this work, we only focus on the deep learning models of video analytics.

\subsection{EC-based Distributed Deep Learning Model}
\begin{figure}[!ht]
  \centering
  \includegraphics[width=3.4in]{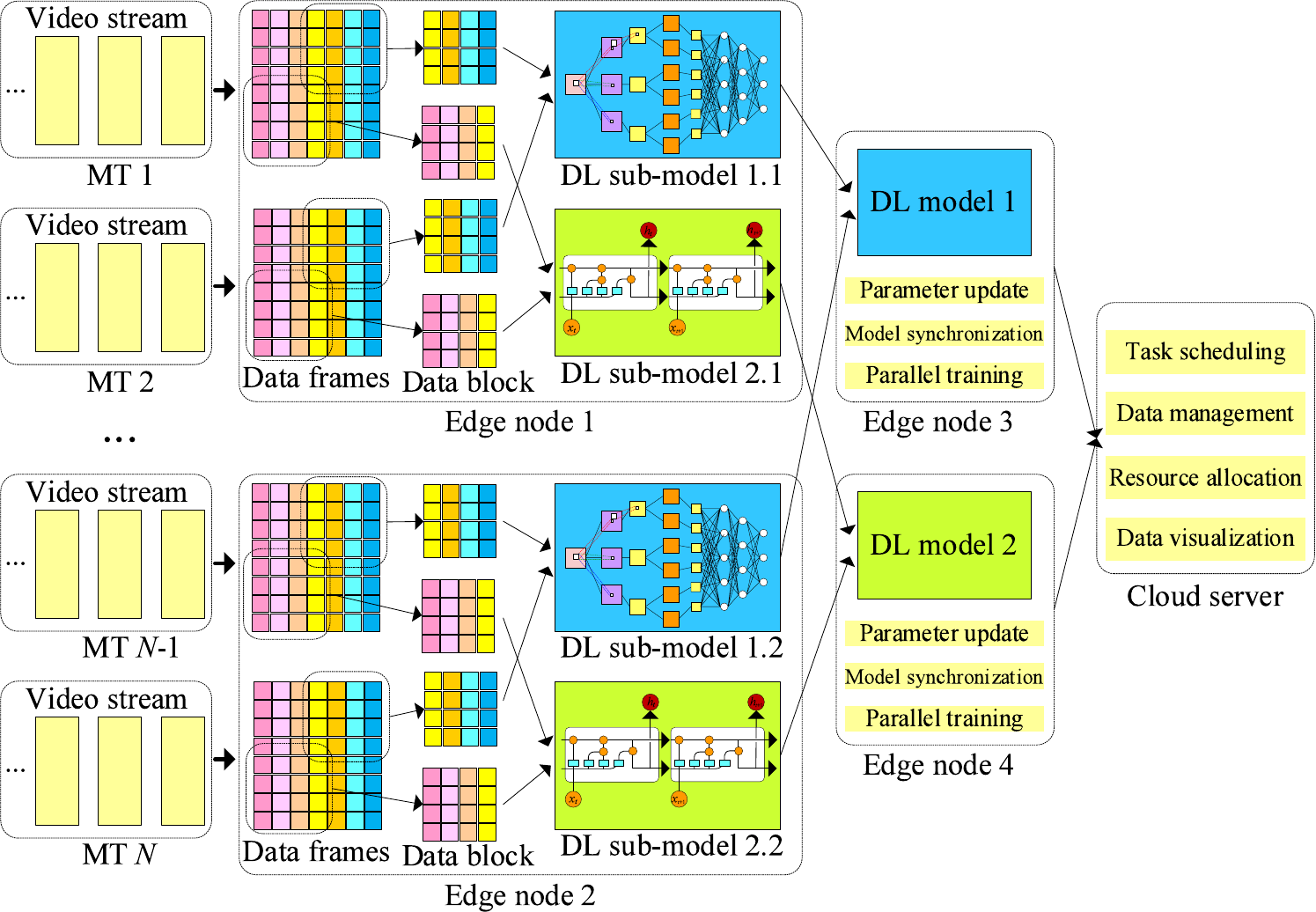}
  \caption{Distributed deep learning model based on edge computing.}
  \label{fig02}
\end{figure}

Most distributed deep learning solutions for big data applications are deployed on distributed cloud centers, leading to a large amount of network communication overheads.
In this work, we propose an EC-based distributed deep learning model for the DIVS system and deploy distributed DL models on the edge nodes.
The proposed distributed deep learning model based on edge computing is illustrated in Fig. \ref{fig02}.

The cloud server is responsible for task scheduling, data management, resource allocation, and data visualization.
During the model training process, the cloud server monitors the training time costs on edge nodes and migrates datasets to achieve workload balancing.

As an input data source, each monitoring terminal submits surveillance video data to the corresponding edge node in a streaming manner.
On the edge node, the video stream from each MT is divided into data frames in a sliding-time-window approach.
Then, according to the data access requirements of separate DL sub-models, data blocks are further extracted from the data frames as an input of each DL sub-model.
After obtaining the input dataset, each DL sub-model trains itself and updates the local weight parameters.
The training process on each edge node is performed in parallel.

\section{Implementation of DIVS System}
Based on the proposed the DIVS system architecture, we implement the DIVS system and address the problems of parallel training, model synchronization, and workload balancing.
We introduce task-level parallel and model-level parallel training methods in Section \ref{section4.1} to further accelerate the video analysis process.
Section \ref{section4.2} presents a model parameter updating method to achieve model synchronization of the global
DL model in a distributed EC environment.
Section \ref{section4.3} presents a dynamic data migration approach is proposed to address the imbalance of workload of edge nodes.

\subsection{Parallel Training of Distributed DL Model}
\label{section4.1}
Benefitting from the formidable and scalable computing capacities of edge nodes, we propose two parallel training approaches of DL training model on the EC environment to accelerate the video analysis process of the DIVS system.

\subsubsection{Task-level Parallel Training}
In actual scenarios, multiple video analytics tasks are typically performed in a video file on a DIVS system.
For example, many researchers applied different deep learning algorithms for traffic monitoring, such as CNN models for vehicle classification \cite{ai02, ec04} and LSTM models for traffic flow prediction \cite{ec06, ec07}.
Therefore, we propose a task-level parallel training method for the distributed DL model.
We deploy multiple DL models with different structures (i.e., CNN and LSTM) on each edge node to perform different data analysis tasks in parallel.
Each DL model is divided into multiple sub-models and allocated to corresponding edge nodes.
Taking the field of traffic monitoring as an example, we deploy two DL models to perform three video surveillance tasks, including a CNN model for vehicle classification and a LSTM model for traffic flow prediction.

(1) CNN model for vehicle classification.

\begin{figure}[!ht]
 \setlength{\abovecaptionskip}{0pt}
 \setlength{\belowcaptionskip}{0pt}
 \centering
 \includegraphics[width=3.45in]{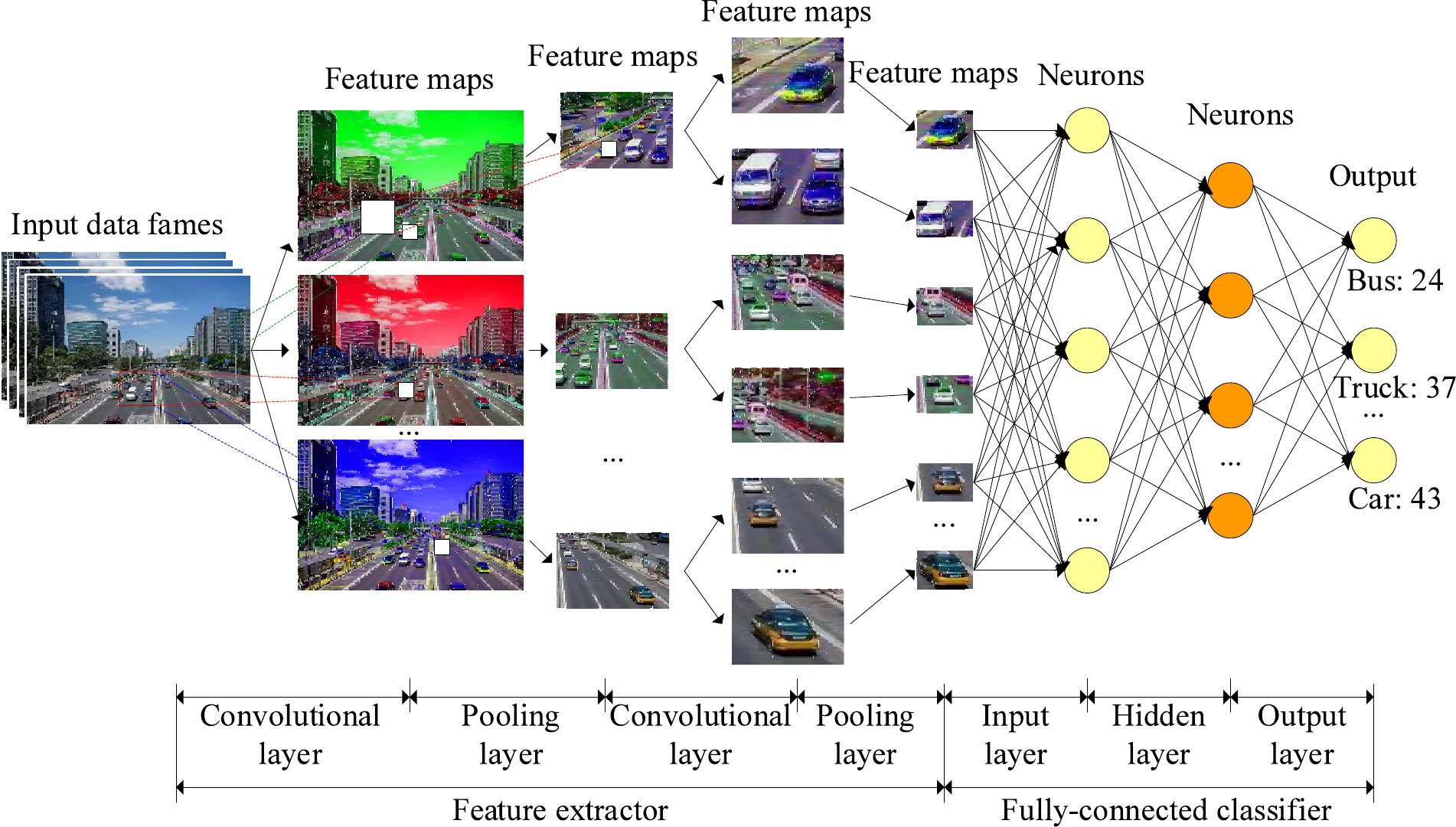}
 \caption{Example of the CNN model for vehicle classification.}
 \label{fig03}
\end{figure}

In existing work, CNN model was widely applied in vehicle classification \cite{ec06, ec07}.
To classify all types of vehicle from the traffic monitoring video streams, we build a distributed CNN model in the DIVS system based on the existing work.
The original traffic monitoring video stream is submitted from each MT to EN and is divided into multiple video frames.
Then, the CNN model extracts all vehicles from each video frame with realistic and complex background and saves them as separate sub-images.
An example of the CNN model structure for vehicle classification is illustrated in Fig. \ref{fig03}.

The structure of CNN vehicle classification model is formulated as flowing:
\begin{center}
{\scriptsize
$INPUT \rightarrow \left\{[L_{CONV}] -> [L_{POOL}] \right\} \times 2 \rightarrow [L_{FC}] \times 3$.
}
\end{center}
In each feature extractor unit, there are one convolutional layer $L_{CONV}$ and one pooling layers $L_{POOL}$.
Repeating this unit 2 time superimpose 3 fully-connected layers $L_{FC}$.

In the first feature extractor unit, the original image of each video frame is sent to a convolutional layer to obtain a feature map.
We use a pooling layer to compress features and determine if the current area contains vehicles.
In the second feature extractor unit, intermediate images containing vehicles are sent to the second convolutional layer to obtain a fine-grained feature map.
From the second pooling layer, we can extract each potential vehicle and sent it as a separate input to the fully-connected layers.
SVM classifiers and Bayesian networks are applied in the fully-connected layers to accurately classify all types of the vehicles.
Finally, the CNN vehicle classification model is copied and distributed into each edge node for parallel training.
Due to limited space, we do not discuss the detail process of the CNN model.

(2) LSTM model for traffic flow prediction.

\begin{figure}[!ht]
 \setlength{\abovecaptionskip}{0pt}
 \setlength{\belowcaptionskip}{0pt}
 \centering
 \includegraphics[width=3.45in]{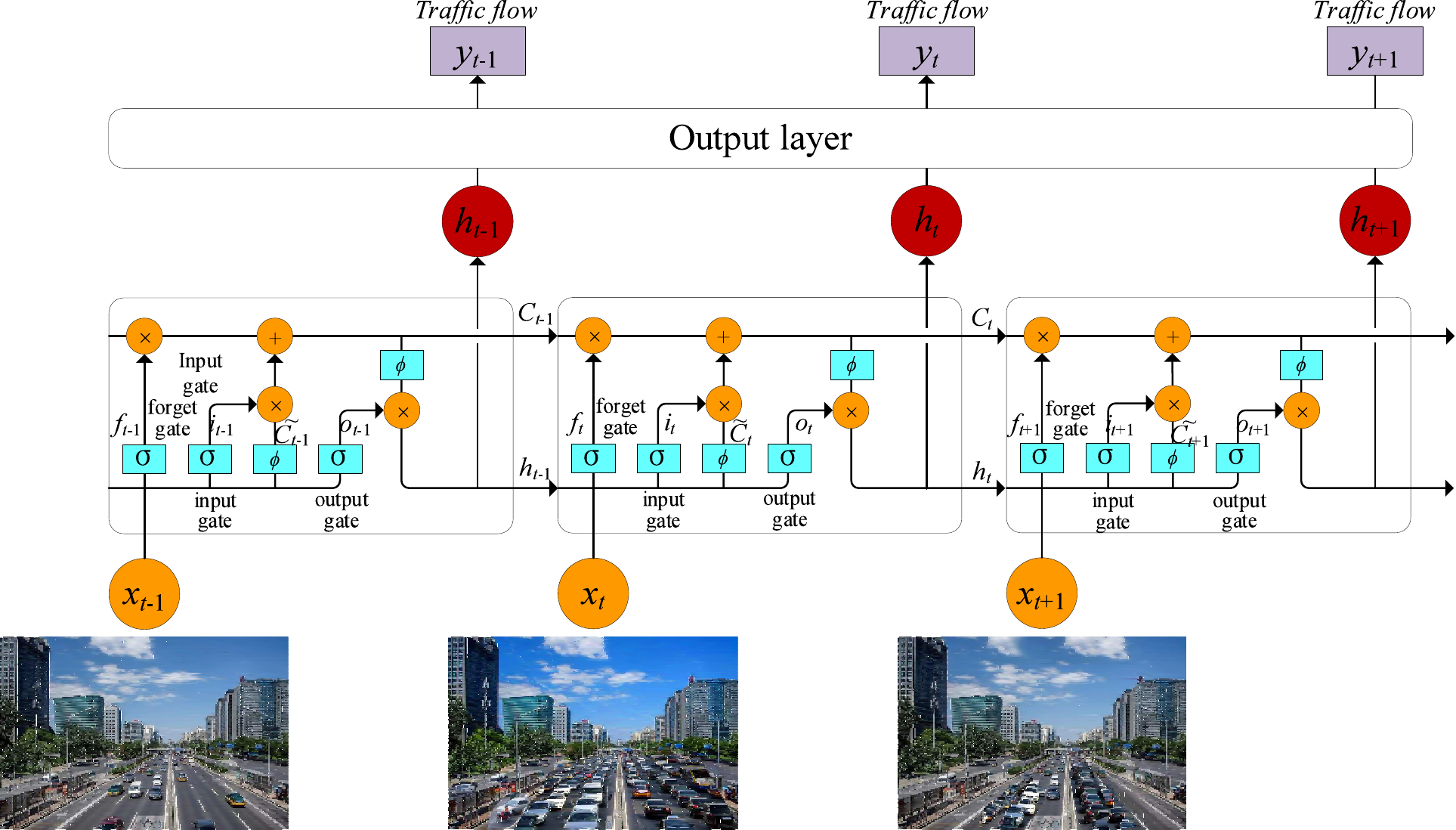}
 \caption{Example of the LSTM model for traffic flow prediction.}
 \label{fig04}
\end{figure}

In existing work, LSTM model was applied in video surveillance, especially for traffic flow prediction \cite{ec06, ec07}.
To make full use of the traffic monitoring video stream, we build a distributed LSTM model based on the existing work for the DIVS system to predict the traffic flow by sharing the same input video stream of the CNN vehicle classification model.
The LSTM model learn the time series with dependencies and automatically evaluates the optimal time periods for time series prediction.
An example of the LSTM model structure for traffic flow prediction is illustrated in Fig. \ref{fig04}.

In this work, the LSTM traffic flow prediction model is composed of an input layer, an recurrent hidden layer, and an output layer.
We express the monitoring video streams as time series, denoted as $X = (X_{1}, ..., X_{t}, ..., X_{T})$, where $x_{t}$ is the video frame at the $t$-th time step.
We calculate the memory cell $C_{t}$ based on the input $X_{t}$, which is a core module of the LSTM model.
Three gates are employed to control the value of $C_{t}$: a forget gate $f_{t}$ to forget the current value of $C_{t}$, an input gate $i_{t}$ to read its input, and an output gate $o_{t}$ to output the value of $C_{t}$.
The gates and cells of the LSTM model are defined as follows:
\begin{equation}
\label{eq01}
\begin{aligned}
i_{t} = & \sigma \left(W_{i} [X_{t}, h_{t-1}] + b_{i}\right),\\
f_{t} = & \sigma \left(W_{f} \times [X_{t},h_{t-1}] + b_{f}\right),\\
o_{t} = & \sigma \left(W_{xo} X_{t} + W_{ho}h_{t-1}\right),\\
\widetilde{C}_{t} = & \phi \left(W_{C} [X_{t},h_{t-1}] + b_{C}\right),\\
C_{t} =& f_{t} \times c_{t-1} + i_{t} \times \widetilde{C}_{t},\\
\end{aligned}
\end{equation}
where $W$ matrices are the weight parameters of the LSTM model.
The LSTM traffic flow prediction model is copied and distributed into each edge node for parallel training.
Due to limited space, we do not discuss the detail process of the LSTM model.
Readers can explore more interesting DL models on each edge node to training the monitoring video datasets.

\subsubsection{Model-level Parallel Training}

Considering that the edge nodes are equipped with multi-core CPUs and have potential parallel computing power, we propose a model-level parallel training method to further accelerate the training process of each DL sub-model on each edge node.
In this section, we use the CNN model as an example to introduce the parallelization of two important training processes, such as the convolutional layers and fully-connected layers.

(1) Parallelization of convolutional layer.

\begin{figure}[ht]
 \setlength{\abovecaptionskip}{0pt}
 \setlength{\belowcaptionskip}{0pt}
 \centering
 \includegraphics[width=2.8in]{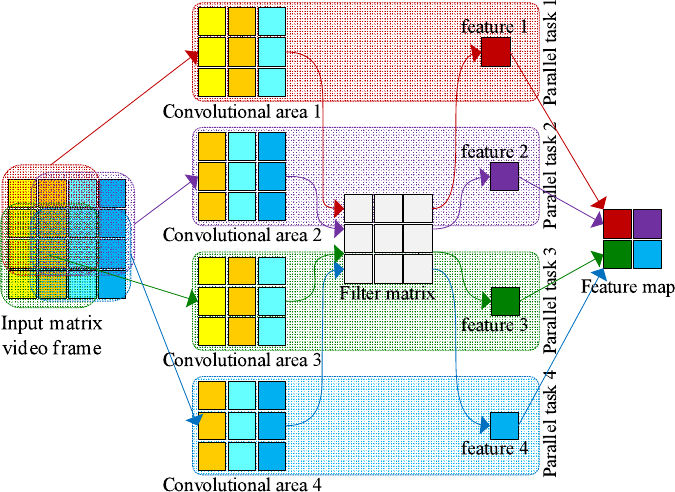}
 \caption{Example of the parallel convolutional operation.}
 \label{fig05}
\end{figure}

A video frame is used as an input matrix $X$ of the CNN sub-model.
In the convolution layer, a filter parameter matrix $F$ is introduced to transfer the input matrix $X$ into a feature map $A$ to extract the key features.
We can partition $X$ into multiple convolution areas to perform the convolutional operation in parallel.
Assuming that ($D_{x}, H_{x}, W_{x}$) is the shape (i.e., depth, height, and width) of $X$ and ($D_{f}, H_{f}, W_{f}$) is the shape of $F$, we can calculate the shape ($D_{a}, H_{a}, W_{a}$) of $A$ as:
\begin{equation}
\label{eq02}
\begin{aligned}
D_{a} &= D_{x} = D_{f},\\
H_{a} &= \frac{H_{x} - H_{f} + 2P}{S} +1,\\
W_{a} &= \frac{W_{x} - W_{f} + 2P}{S} +1,
\end{aligned}
\end{equation}
where $P$ is the number of the zero padding of $X$ and $S$ is the stride of the convolutional operation.
Then, we extract each convolutional area $X[r_{s} : r_{e}, ~c_{s}: c_{e}]$ (the start and end rows and columns) of $X$ for each parallel task:
\begin{equation}
\label{eq03}
\begin{aligned}
r_{s} &= i \times S, &c_{s} &= j \times S, \\
r_{e} &= r_{s} + H_{f}, &c_{e} &= c_{s} + W_{f}.
\end{aligned}
\end{equation}
Each convolution area is convoluted separately with the filter parameter matrix to get the result: an element of the feature map.
Each element in the feature map $A$ is computed based on the corresponding convolution area in $X$ and $F$.
Different tasks can access different convolution areas in $X$ simultaneously without updating the their values, there is not any data dependency among these tasks.
An example of the parallel convolutional computation of each CNN sub-model is illustrated in Fig. \ref{fig05} and the steps of this process are described in Algorithm \ref{alg04}.

\begin{algorithm}[!ht]
\caption{{\scriptsize Parallel convolutional operation of distributed CNN model.}}
\scriptsize
\label{alg04}
\begin{algorithmic}[1]
\REQUIRE ~\\
$X$: The input matrix of the training vide frames;\\
$F$: the filter parameter matrix of the current CNN sub-model.\\
\ENSURE ~\\
$A$: the feature map of the current CNN sub-model.\\
\STATE compute the shape ($D_{a}$, $H_{a}$, $W_{a}$) of $A$ based on $X$ and $F$;
\STATE compute the number of parallel tasks $P_{Conv} = D_{a} \times H_{a} \times W_{a}$;
\FOR {each parallel task $T_{p}$ in $P_{Conv}$}
\STATE  extract convolutional area $X_{i} \leftarrow X[r_{s} : r_{e}, ~c_{s}: c_{e}]$ from $X$;
\STATE  execute convolutional operation $a_{i} \leftarrow Conv(X_{i} F)$;
\STATE  append to the feature map $A \leftarrow a_{i}$;
\ENDFOR
\RETURN $A$.
\end{algorithmic}
\end{algorithm}

(2) Parallelization of fully-connected layer.

In a fully-connected network, different neurons are arranged in different layers, with multiple neurons in each layer.
Each neuron in the $i$-th layer $L_{i}$ is connected to all the neurons in $L_{i-1}$, and the output of neurons in $L_{i-1}$ is the input of neurons in $L_{i}$.
That is, data dependencies occur among neurons in $L_{i}$ and $L_{i-1}$.
In contrast, there is no connection among neurons in the same layer, namely, there is no logical or data dependency between them.
Therefore, the calculation process of neurons in the same layer can be executed in parallel.

\begin{figure}[!ht]
  \centering
  \subfigure[Tasks of a hidden layer]{\includegraphics[width=1.5in]{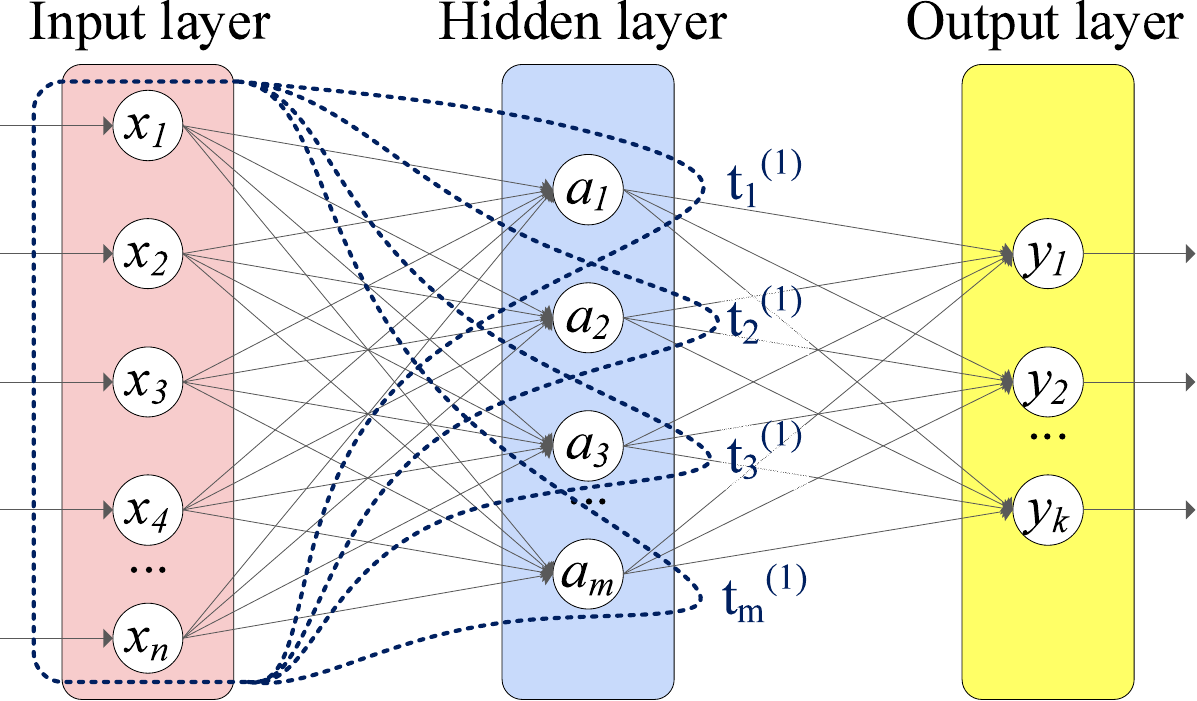}}
  \subfigure[Tasks of the output layer]{\includegraphics[width=1.7in]{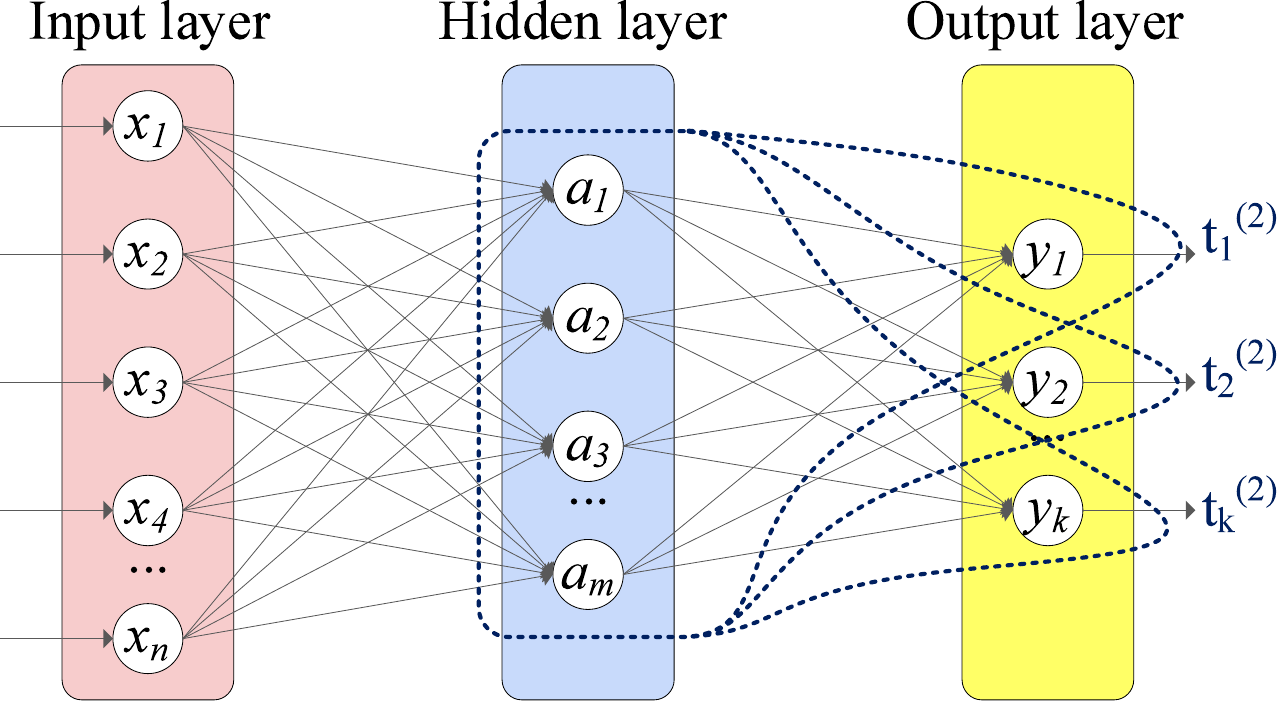}}
  \caption{Parallel training process of the fully-connected network.}
  \label{fig06}
\end{figure}

Assuming that there are $n_{i-1}$ neurons in the layer $L_{i-1}$ and $n_{i}$ neurons in $L_{i}$, and $W_{i-1,i}$ is the weight set for the neurons in $L_{i-1}$ and $L_{i}$.
For each neuron $a_{ij}$ in $L_{i}$, we calculate the output of $a_{ij}$, as defined in Eq. (\ref{eq04}):
\begin{equation}
\label{eq04}
a_{ij} = f(\sum_{j=1}^{n_{i-1}}{w_{ji} x_{j}} + w_{b}),
\end{equation}
where $f()$ is the activation function between $L_{i-1}$ and $L_{i}$.
The computation tasks of each neuron $a_{ij}$ in $L_{i}$ can be executed in parallel.
An example of parallel training of the fully-connected network is illustrated in Fig. \ref{fig06}.

In Fig. \ref{fig06} (a), we parallelize the computation tasks of each hidden layer into $m$ independent sub-tasks $\{t_{1}^{(1)}, t_{2}^{(1)}, ..., t_{m}^{(1)}\}$.
In addition, after the outputs of all neurons in the hidden layer are obtained, we also calculated each neuron in the output layer simultaneously.
As shown in Fig. \ref{fig06} (b) The computation tasks of the output layer are decomposed into $k$ parallelizable sub-tasks $\{t_{1}^{(2)}, t_{2}^{(2)}, ..., t_{k}^{(2)}\}$.

The maximum parallelism degree of the entire fully-connected network is the neuron width of the entire fully-connected layer, namely, the number of neurons in the layer having the largest number of neurons, as defined in Eq. (\ref{eq05}):
\begin{equation}
\label{eq05}
\hbar_{FC} = \max_{i=1}^{k}{|n_{i}|},
\end{equation}
where $|n_{i}|$ is the number of neuron in the layer $L_{i}$.

\subsection{Weight Parameter Update and Model Synchronization}
\label{section4.2}
\begin{figure}[!ht]
  \centering
  \includegraphics[width=3.4in]{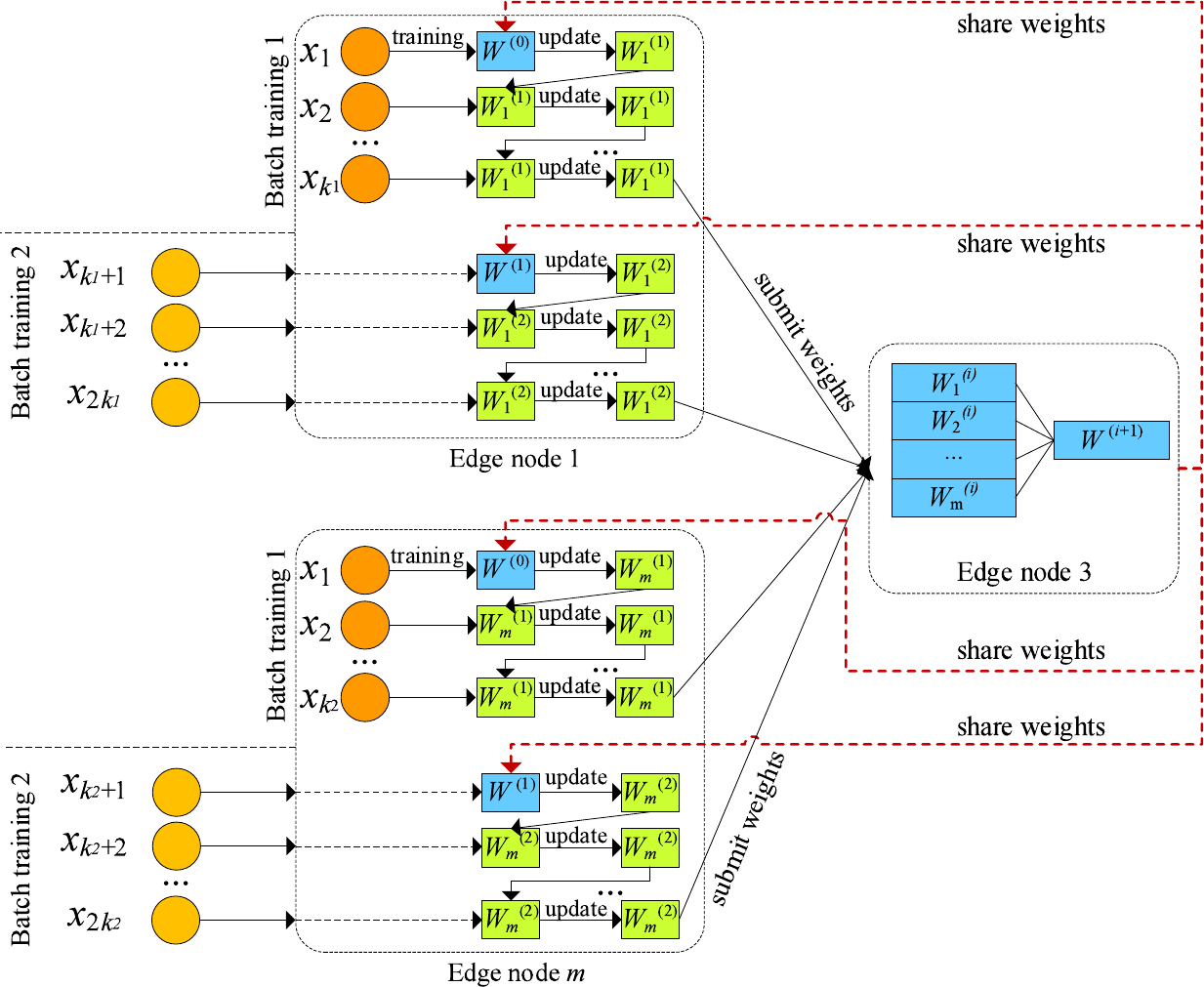}
  \caption{Workflow of weight parameter update and model synchronization.}
  \label{fig07}
\end{figure}

In actual scenarios, each edge node may connect to a different number of monitoring terminals due to uneven deployment of monitoring terminals.
In addition, each edge node has different computing capacities due to the heterogeneity of edge nodes.
These conditions lead to different workloads and training speeds between edge nodes of the same level, and further lead to synchronous problems during global weight updating.
Therefore, we propose a weight parameter update and model synchronization method for the distributed DL model.
The workflow of the proposed weight parameter update and model synchronization method is illustrated in Fig. \ref{fig07}.

\textbf{Definition 1: Local weight set}.
\textit{The weight parameters of each DL sub-model on each edge node is defined as a local weight set.
Each local weight set is trained based on the local video frames and updated on a low-level edge node.}

\textbf{Definition 2: Global weight set}.
\textit{The weight parameters of the entire DL model is defined as a global weight set.
The global weight set is updated on a high-level edge node by collecting the local weight sets of all of its sub-models.}

We use a batch training approach to train the DL sub-model and update the corresponding local weight set.
As shown in Fig. \ref{fig07}, the initial global weight set $W^{(0)}$ of a DL model is shared to all low-level ENs to train the corresponding sub-models.
All edge nodes use $W^{(0)}$ to train the first sample of the first batch and get the corresponding outputs.
Each edge node $EN_{j}$ updates $W^{(0)}$ as a local weight set $W_{j}^{(1)}$ based on the output.
Then, $W_{j}^{(1)}$ is used to train the second sample on $EN_{j}$ and obtain a updated value of $W_{j}^{(1)}$.
Repeat this step, until all of the samples in the first batch on $EN_{j}$ are trained.

After all edge nodes complete a batch training, the latest local weight set trained on each edge node is aggregated to a high-level edge node to calculate a new version of global weight set $W^{(1)}$.
Given $N$ MTs and $M$ ENs in a DIVS system, the average number of MTs connected to each EN is set as $\overline{n} = N/M$.
Assuming that for each MT, the video stream per unit time is divided into $\alpha$ video frames.
The video frames on each edge node are further divided into multiple batches of training subsets, let $b$ be the number of batches.
Assume that there are $N_{j}$ video streams from $N_{j}$ MTs arrive to the $j$-th edge node $EN_{j}$, the batch size on $EN_{j}$ is calculated as $k_{j} =\frac{N_{j}\times \alpha}{b}$, and the average number of samples in each training batch on an edge node is $\overline{k} =\frac{N \times \alpha}{M \times b}$.
Hence, we define the exponential value of the difference between the batch width of each edge node and the average batch width of all edge nodes as its contribution to the global weight set:
\begin{equation}
\label{eq06}
Q_{j} = e^{(k_{j} - \overline{k})} =e^{\left(\frac{N_{j}\times \alpha}{b} - \frac{N \times \alpha}{M \times b}\right)}.
\end{equation}
The global weight set $W^{(i)}$ for the $(i)$-th batch training is defined:
\begin{equation}
\label{eq07}
W^{(i)} = \sum_{j=1}^{m}{\left(W_{j}^{(i)} \times Q_{j}\right)},
\end{equation}
where $W_{j}^{(i)}$ is the updated local weight set the DL sub-model on $EN_{j}$, which is trained by the $(i-1)$-th batch training.
After obtaining the updated global weight set $W^{(i)}$, the entire DL model achieves a synchronization.
Then, the high-level edge node shares $W^{(i)}$ to each edge node for the next batch training.
The above process is repeated until the weight set reaches a steady state.
The steps of weight parameter update and model synchronization is described in Algorithm \ref{alg02}.

\begin{algorithm}[!ht]
\scriptsize
\caption{{\scriptsize Weight parameter update and model synchronization.}}
\label{alg02}
\begin{algorithmic}[1]
\REQUIRE ~\\
$W[i]$: the list of local weight sets trained based on $W^{(i-1)}$, $W[i] = \{W_{1}^{(i)}, ..., W_{m}^{(i)}\}$;\\
$Q$: the list of batch sizes of all edge nodes, $Q = \{Q_{1}, ..., Q_{m}\}$.\\
\ENSURE ~\\
$W^{(i)}$: the new version of the global weight set.\\
\FOR {each local weight set $W_{j}^{(i)}$ in $W[i]$}
\FOR {each weight parameter $w_{j,\kappa}$ in $W_{j}^{(i)}$}
\STATE calculate global weight parameter $w_{\kappa}^{(i)}$ $\leftarrow$ $w_{\kappa}^{(i)} + w_{j,\kappa} \times Q_{j}$;
\STATE append to the global weight set $W^{(i)}$ $\leftarrow$ $w_{\kappa}^{(i)}$;
\ENDFOR
\ENDFOR
\RETURN $W^{(i)}$.
\end{algorithmic}
\end{algorithm}

\subsection{Dynamic Video Data Migration}
\label{section4.3}
As described in previous section, edge nodes have different workloads and training speeds, which will lead to synchronous problems during global weight updating.
Therefore, we propose a Dynamic Data Migration (DDM) strategy to maximize workload balancing of the distributed EC system and minimize the synchronous in global weight updating.

Let $ENs = \{EN_{1}, ..., EN_{M}\}$ be $M$ edge nodes in the DIVS system.
Assuming that there are $n_{j}$ video frames in the training subset on computer $EN_{j}$, and $\overline{t_{j}}$ is the average time it takes for $EN_{j}$ to complete a training process on a video frame in the current migration assessment period.
The migration assessment period is defined as a time period between two migration operations.
Hence, the time of an epoch of iteration training on $EN_{j}$ is calculated as $T_{j} = \overline{t_{j}} \times n_{j}$.

Let $\overline{T}$ be the average time that all edge nodes complete a training iteration.
If $(\overline{T}-T_{j}) < 0$, it means $EN_{j}$ needs to migrate video frames to reduce the training time.
If $(\overline{T}-T_{j}) > 0$, it means $EN_{j}$ has available computing resources to receive some immigrated video frames.
If $(\overline{T}-T_{j}) = 0$, it means $EN_{j}$ does not need to migrate as well as cannot receive immigrated video frames.
The number of video frames on $EN_{j}$ that require to be migrated is calculated in Eq. (\ref{eq08}):
\begin{equation}
\label{eq08}
\triangle n_{j} = \lfloor{\frac{\overline{T}-T_{j}}{\overline{t_{j}}}}\rfloor.
\end{equation}
The set of migration data is denoted as $\triangle N = \{\triangle n_{1}, \triangle n_{2}, ..., \triangle n_{m} \}$.
$\triangle n_{j}< 0$ indicates the number of video frames to be migrated, and set $n_{out} = \triangle n_{j}$.
$\triangle n_{j} >0$ indicates the number of video frames allowed to be immigrated, and set $n_{in} = \triangle n_{j}$.
After migration, the new execution time $T_{j}^{'}$ on $EN_{j}$ id calculated as $T_{j}^{'} = \overline{t_{j}} \times (n_{j} + \triangle n_{j})$.
Therefore, the problem of workload balancing of the EC system is formulated in Eq. (\ref{eq09}):
\begin{equation}
\label{eq09}
\mathbb{B} = arg~ \min(\sqrt{\frac{\sum_{j=1}^{m}{(T_{j}^{'} - \overline{T})^{2}}}{m}}).
\end{equation}
The smaller the value of $\Theta$, the more balanced the workload of the entire EC system.
Let $\theta_{\mathbb{B}}$ be the threshold of the workload balance.
If $\mathbb{B} \geq \theta_{\mathbb{B}}$, we continue to calculate the number of migrations and the migration plan.
Otherwise, no migration is done.
Namely, instead of migrating all the time, we only conduct migration assessments in each migration interval.

After getting the amount of video frames that each edge node requires to migrate, we match the migration requirements and immigration capacities for computers.
We try to move the entire dataset to the target edge nodes to guarantee a minimum number of edge nodes involved in each migration.
We introduce an offset of $\xi$ for the data migration matching.
$|n_{in}| \approx |n_{out}|$, if $|n_{in} +  n_{out}| \leq \xi$, where $n_{in}$ is the number of video frames that allow to move in and and $n_{out}$ is that of video frames that need to move out.
The data migration matching strategy is described as follows:

\begin{enumerate}[label={\bfseries P\arabic*}]
\item Find edge nodes that do not meet the migration conditions.
    For an edge node $EN_{in}$, if $|n_{in}| \leq \xi$, namely, $EN_{in}$ does not need to migrate data and is removed from the migration list.
    Similarly, if $|n_{out}| \leq \xi$, it means that $EN_{out}$ cannot provide a significant resource for data immigration and is removed from the immigration list.
\item Find edge nodes with the best migration matching.
    If $|n_{in}| \approx |n_{out}|$, then the data migration will be performed between edge nodes $EN_{in}$ and $EN_{out}$ with the amount of $min (n_{in}, n_{out})$.
\item Find $max(n_{in})$ from the current immigration list, then find $max(|n_{out}|)$ from the migration list.
    It is easy to prove that $|max(|n_{in}|) - max(|n_{out}|)|> \xi$.
\end{enumerate}

\begin{algorithm}[!ht]
\caption{{\scriptsize Dynamic data migration strategy of the DIVS system.}}
\label{alg03}
\begin{algorithmic}[1]
{\scriptsize \REQUIRE ~\\
$EN_{s}$: edge nodes in the distributed edge computing system;\\
$T_{s}$: training time list of $EN_{s}$ in the current migration period;\\
$\xi$: an offset of the data migration matching.\\
\ENSURE ~\\
$DM_{s}$: the list of data migration matching results.\\
\STATE calculate the average execution time $\overline{T}$ $\leftarrow$ avg($T_{s}$);
\FOR {each edge node $EN_{j}$ in $EN_{s}$}
\STATE calculate migration amount $\triangle n_{j}$ $\leftarrow$ $\lfloor{\frac{\overline{T}-T_{j}}{\overline{t_{j}}}\rfloor}$;
\ENDFOR
\STATE build migration list $L_{Out}$ $\leftarrow$ $EN_{j}|$($\triangle n_{j}<0$ and $|\triangle n_{j}|> \xi$);
\STATE build immigration list $L_{In}$ $\leftarrow$ $EN_{j}|$($\triangle n_{j}>0$ and $|\triangle n_{j}|> \xi$);
\FOR {each edge $EN_{out}$ in $L_{Out}$}
\FOR {each edge $EN_{in}$ in $L_{In}$}
\IF {$|n_{out} + n_{in}| \leq \xi$}
\STATE exec migration $DM_{s}$ $\leftarrow$ [$EN_{out}, EN_{in}, min(n_{in}, n_{out})$];
\STATE remove $EN_{out}$ from $L_{Out}$, $EN_{in}$ from $L_{In}$;
\ENDIF
\ENDFOR
\ENDFOR
\STATE \textbf{while} {$L_{Out} \neq NULL$ and $L_{In} \neq NULL$} \textbf{do}
\STATE \quad find $max(n_{out})$ from $L_{Out}$, $max(n_{in})$ from $L_{In}$;
\STATE \quad \textbf{if} $max(|n_{in}|) > max(|n_{out}|)$ \textbf{do}
\STATE \qquad execute migration $DM_{s}$ $\leftarrow$ [$EN_{out}, EN_{in}, |n_{out}|$];
\STATE \qquad remove $EN_{out}$ from $L_{Out}$, set $n_{in} \leftarrow (n_{in} - |n_{out}|)$;
\STATE \quad \textbf{else}
\STATE \qquad execute migration $DM_{s}$ $\leftarrow$ [$EN_{out}, EN_{in}, |n_{in}|$];
\STATE \qquad remove $EN_{in}$ from $L_{In}$, set $n_{out} \leftarrow (n_{out} + n_{in})$;
\RETURN $DM_{s}$.}
\end{algorithmic}
\end{algorithm}

\section{Experiments}
\label{section5}

\subsection{Experimental Setting}
We conduct experiments to evaluate the effectiveness and efficiency of the proposed EC-based DIVS system.
We built an EC system with two levels of edge nodes, including 200 monitoring terminals, 35 EC servers, and a cloud server.
Monitoring terminals are deployed at the intersections of 30 streets.
Then, 30 first-level edge nodes are deployed on the corresponding streets to collect video streams from MTs in the current street.
There are 5 ENs in the second level of the EC system, each of which is connected to 6 first-level ENs.
Half of the edge nodes are equipped with Intel Core i5-6400 quad-core CPU, 6 GB DRAM and 32 GB main memory.
The remaining edge nodes are equipped with Intel Xeon Nehalem EX six-core CPU, 8 GB DRAM and 64 GB main memory.
A high-speed Gigabit network is used between MTs and the edge nodes.
In the experiments, seven days of traffic monitoring videos are collected from these monitoring terminals.

\subsection{Performance Evaluation}
\begin{figure}[!ht]
 \centering
 \subfigure[Different edge node scales]{\includegraphics[width=1.7in]{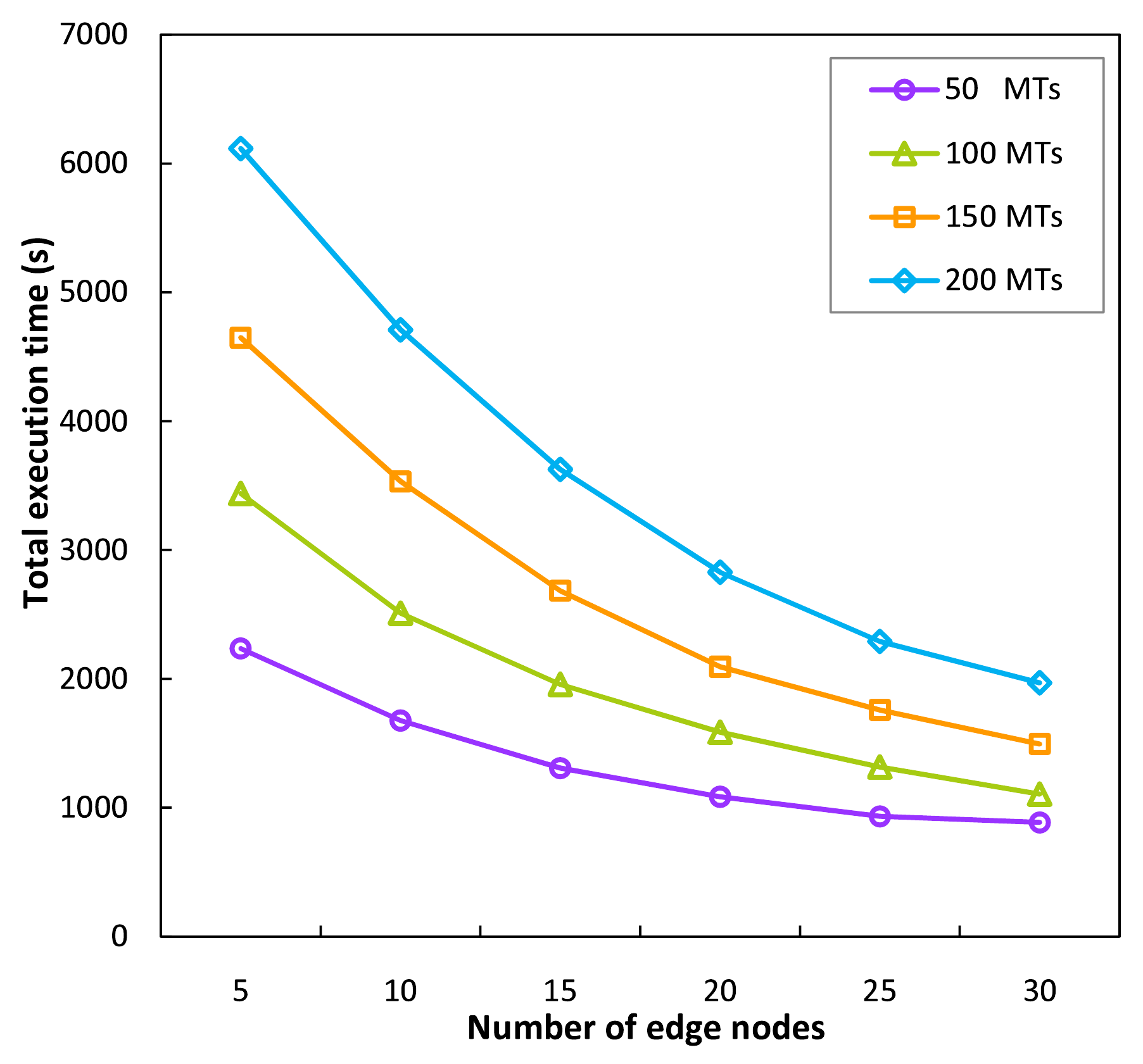}}
 \subfigure[Different task scales]{\includegraphics[width=1.7in]{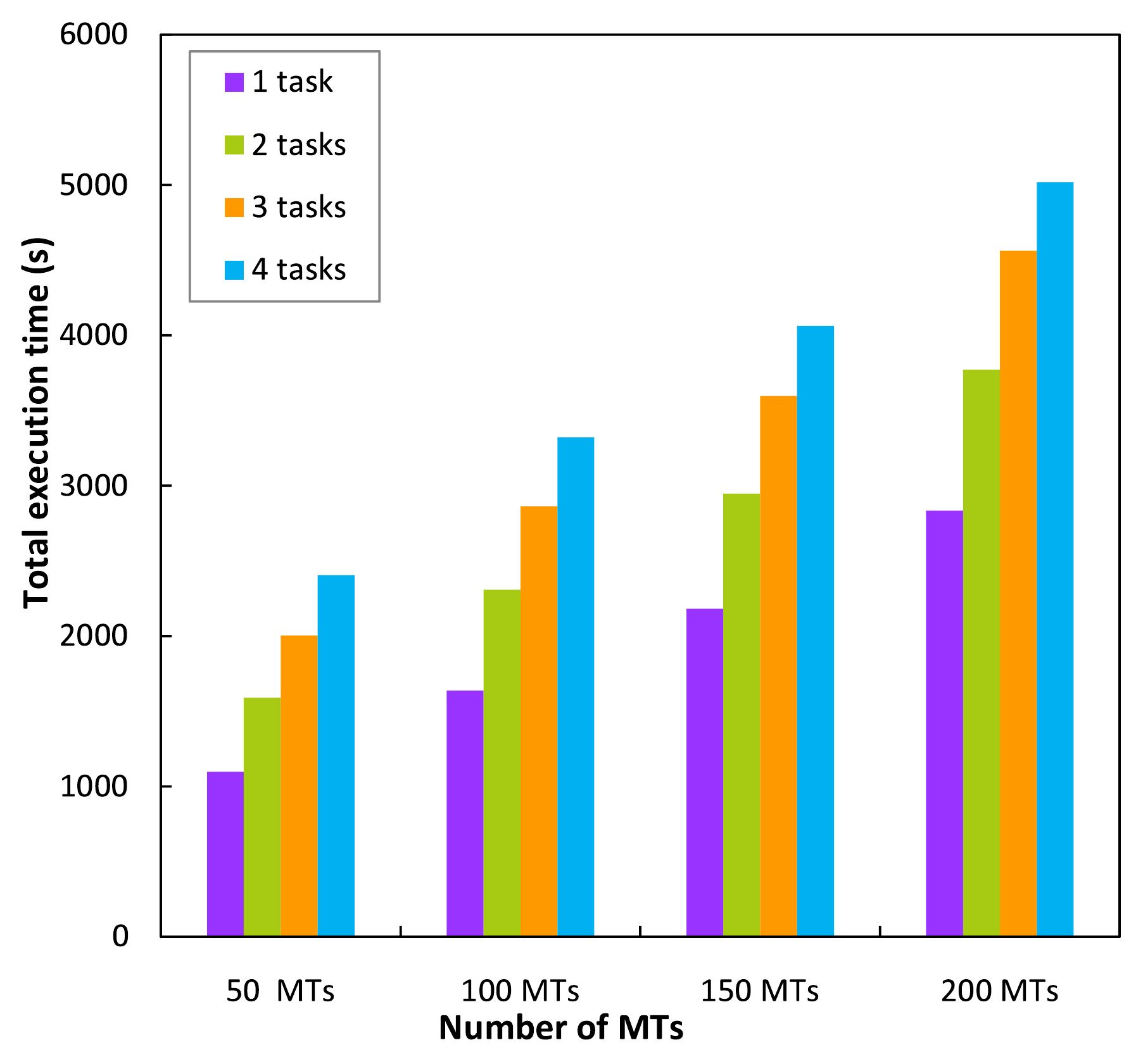}}
 \caption{Impact of edge node and analysis task scales on performance.}
 \label{chart01}
\end{figure}

We evaluate the performance of the proposed DIVS system from the perspective of the scale of edge nodes and video analysis tasks.
To be fair, we set the structure and parameters of the DL models so that each DL model has similar computational complexity.
The total execution time of the entire DIVS system is recorded and compared, as shown in Fig. \ref{chart01}.

In each case in Fig. \ref{chart01} (a) with the same number of MTs, as the number of edge nodes increases, the total execution time of the DIVS system continues to decrease.
When the number of edge nodes increases to 30, it only needs 1773.62 (s) and 851.28 (s) respectively.
Interestingly, the advantages in large-scale MT scenarios are more obvious than in small-scale MT scenarios.
For example, when the system is expanded from 5 edge nodes to 30 edge nodes, the execution time of 50 MTs drops from 2335.13 (s) to 851.28 (s), and the decline rate is 61.91\%.
In contrast, the execution time of 200 MTs is reduced from 6117.22 (s) to 1773.62 (s), with a decline rate of 71.01\%.
It is obvious in Fig. \ref{chart01} (b) that an increase in the scale of tasks will not lead to a doubling of the total execution time.
The reason is that no matter how many tasks are required on each edge node, the input video stream needs to be transmitted from the MTs to the edge nodes only once, which saves considerable data communication delay.
In addition, the advantages of the execution time for multiple tasks are more pronounced as the number of MTs increases.
Therefore, the experimental results show that the DIVS system achieves good scalability.

\subsection{Data Communication and Workload Balance}

\begin{figure}[!ht]
 \centering
 \subfigure[Impact of MT scale]{\includegraphics[width=1.7in]{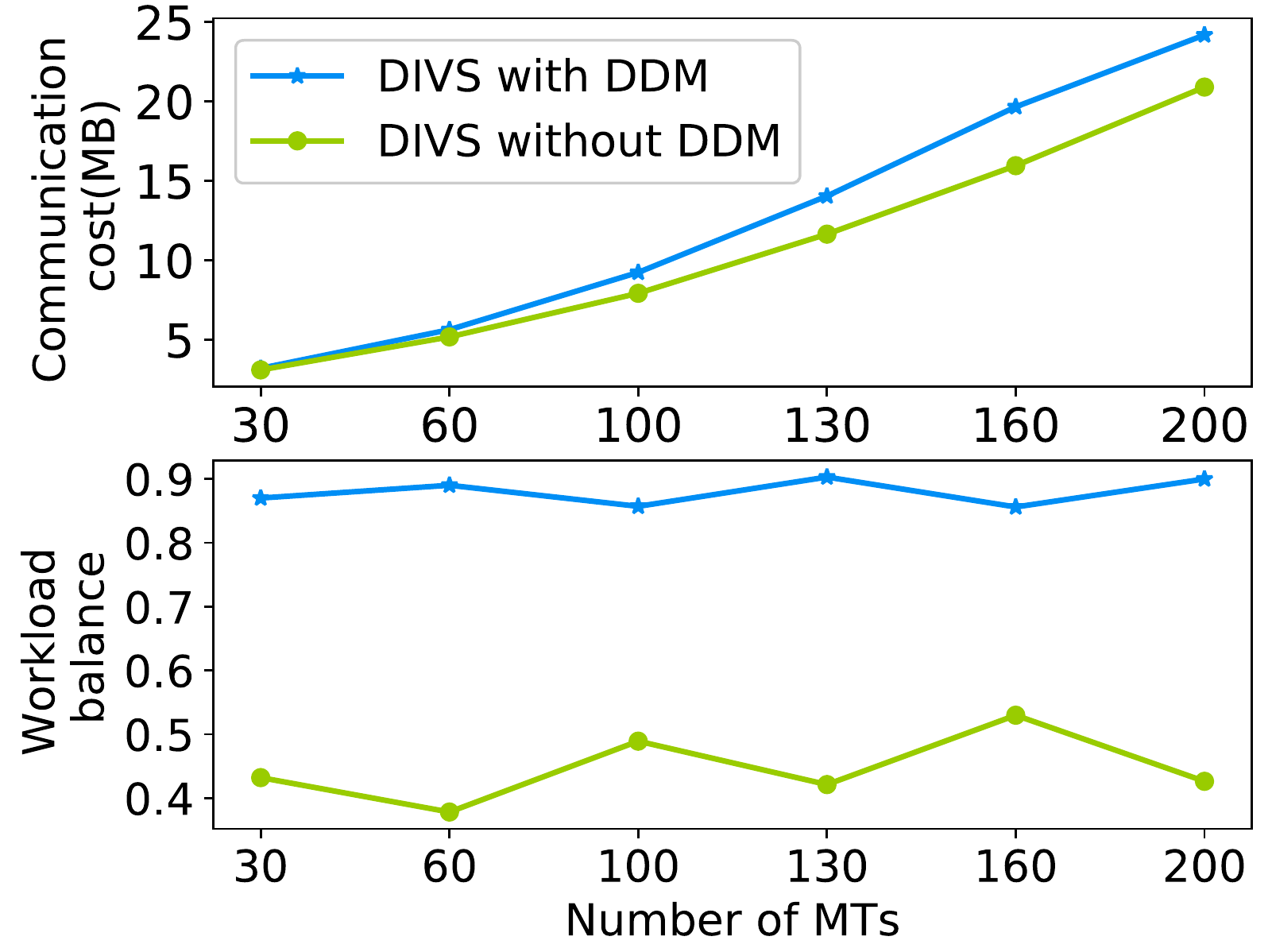}}
 \subfigure[Impact of edge node scale]{\includegraphics[width=1.7in]{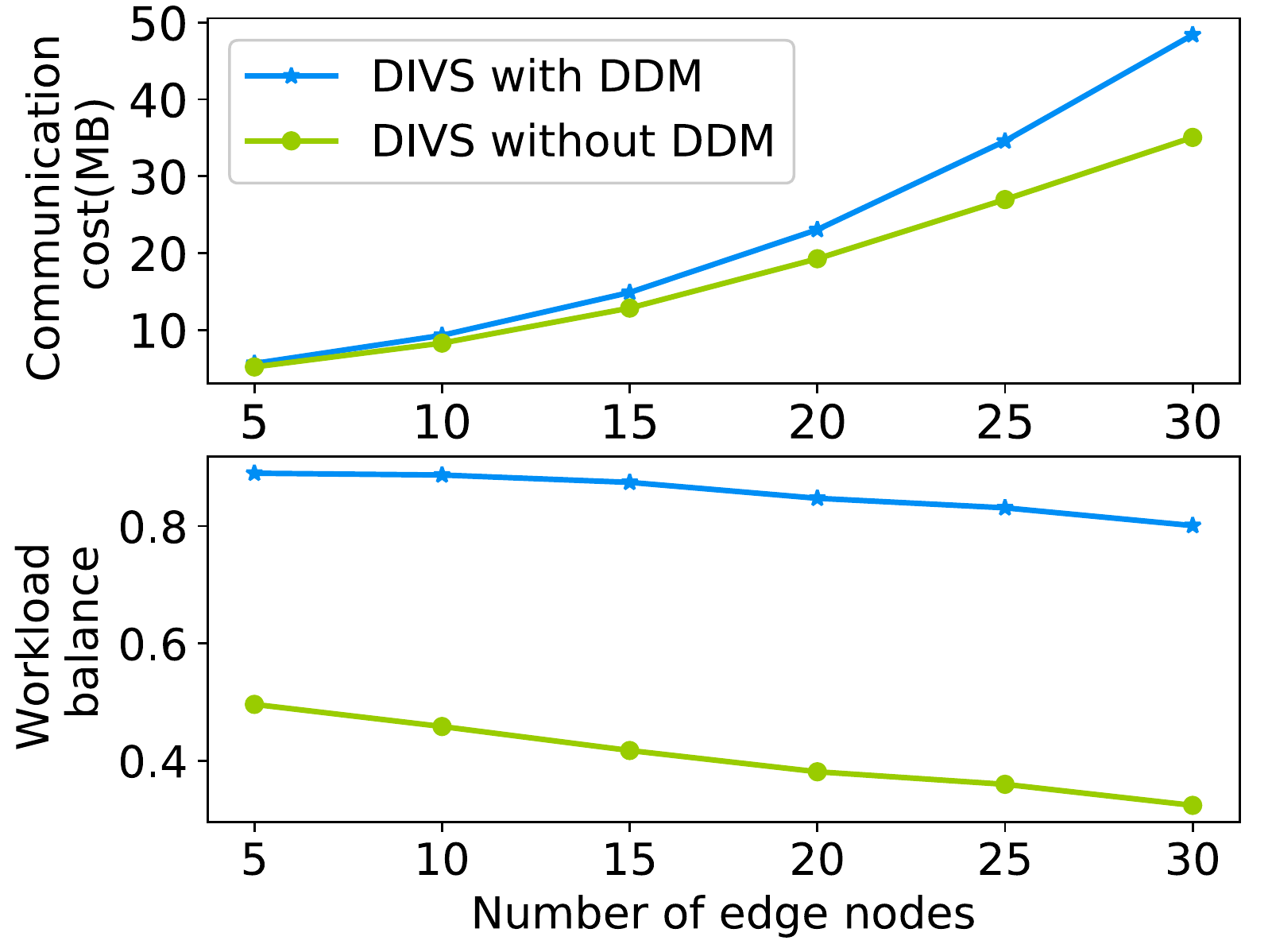}}
 \caption{Data communication and workload balance of DIVS.}
 \label{chart02}
\end{figure}

We evaluate the proposed DIVS architecture in the view of data communication cost and workload balance.
The number of MTs gradually increases from 30 to 200, and the number of edge nodes increases from 5 to 30 in each case.
The experimental results are shown in Fig. \ref{chart02}.

It is clear from Fig. \ref{chart02} that the DDM strategy owns the most significant workload balance with a compromise data communication cost in most cases.
As shown in the curves of DDM in Fig. \ref{chart02} (a), benefitting from dynamic data migration, workload balance of the EC system keeps well steady with the scale of MTs increases.
In contrast, without DDM strategy, although they own lower data communication cost in most cases, but the workload on the edge nodes is seriously unbalanced, which also leads to long waiting time for synchronization and more execution time for the entire DIVS system.
In addition, with the EC system scale increases in Fig. \ref{chart02} (b), benefitting from dynamic data migration, workload balance of the EC system also keeps well steady in the case of DDM.
Experimental results demonstrate that the DDM strategy of DIVS significantly improve the workload balance of the EC system with acceptable communication cost.

\section{Conclusions}
In this paper, we built a Distributed Intelligent Video Surveillance (DIVS) system and deployed it on a multi-layer edge computing architecture to provide flexible and scalable training capabilities.
We addressed the problems of parallel training, model synchronization, and workload balancing.
Two parallelization aspects were provided to improve the throughout of the DIVS system and a model parameter updating method was proposed to realize the model synchronization of the global DL model.
In addition, we proposed a dynamic data migration approach to address the workload and computational power imbalance problems of edge nodes.
Experimental results showed that the DIVS system can efficiently handle video surveillance and analysis tasks.

\section*{Acknowledgment}
This work is partially funded by the National Key R\&D Program of China (Grant No. 2018YFB1003401),
the National Outstanding Youth Science Program of National Natural Science Foundation of China (Grant No. 61625202),
the International (Regional) Cooperation and Exchange Program of National Natural Science Foundation of China (Grant No. 61661146006, 61860206011),
and the International Postdoctoral Exchange Fellowship Program (Grant No. 2018024).
This work is also supported in part by NSF through grants IIS-1526499, IIS-1763325, CNS-1626432, and NSFC 61672313.

\ifCLASSOPTIONcaptionsoff
\newpage
\fi
\bibliographystyle{IEEEtran}
\bibliography{reference}

\begin{thebibliography}{10}
\providecommand{\url}[1]{#1}
\csname url@samestyle\endcsname
\providecommand{\newblock}{\relax}
\providecommand{\bibinfo}[2]{#2}
\providecommand{\BIBentrySTDinterwordspacing}{\spaceskip=0pt\relax}
\providecommand{\BIBentryALTinterwordstretchfactor}{4}
\providecommand{\BIBentryALTinterwordspacing}{\spaceskip=\fontdimen2\font plus
\BIBentryALTinterwordstretchfactor\fontdimen3\font minus
  \fontdimen4\font\relax}
\providecommand{\BIBforeignlanguage}[2]{{%
\expandafter\ifx\csname l@#1\endcsname\relax
\typeout{** WARNING: IEEEtran.bst: No hyphenation pattern has been}%
\typeout{** loaded for the language `#1'. Using the pattern for}%
\typeout{** the default language instead.}%
\else
\language=\csname l@#1\endcsname
\fi
#2}}
\providecommand{\BIBdecl}{\relax}
\BIBdecl

\bibitem{tii05}
U.~L.~N. Puvvadi, K.~D. Benedetto, A.~Patil, K.-D. Kang, and Y.~Park,
  ``Cost-effective security support in real-time video surveillance,''
  \emph{IEEE Trans Ind. Informat.}, vol.~11, no.~6, pp. 1457--1465, 2015.

\bibitem{ai06}
Y.~Zhou, L.~Liu, L.~Shao, and M.~Mellor, ``Fast automatic vehicle annotation
  for urban traffic surveillance,'' \emph{IEEE Trans. Intell. Transport.
  Syst.}, vol.~19, no.~6, pp. 1973--1984, 2018.

\bibitem{vs04}
M.~Valera, S.~A. Velastin, A.~Ellis, and J.~Ferryman, ``Communication
  mechanisms and middleware for distributed video surveillance,'' \emph{IEEE
  Trans. Circuits Syst. Video Technol.}, vol.~21, no.~2, pp. 1795--1809, 2011.

\bibitem{ai01}
Z.~Zhou, H.~Liao, B.~Gu, K.~M.~S. Huq, S.~Mumtaz, and J.~Rodriguez, ``Robust
  mobile crowd sensing: When deep learning meets edge computing,'' \emph{IEEE
  Network}, vol.~32, no.~4, pp. 54--60, 2018.

\bibitem{ai02}
C.~Ding and D.~Tao, ``Trunk-branch ensemble convolutional neural networks for
  video-based face recognition,'' \emph{IEEE Trans. Pattern Anal. Machine
  Intell.}, vol.~40, no.~4, pp. 1002--1014, 2018.

\bibitem{ai03}
L.~Ding, Y.~Tian, H.~Fan, Y.~Wang, and T.~Huang, ``Rate-performance-loss
  optimization for inter-frame deep feature coding from videos,'' \emph{IEEE
  Trans. Image Processing}, vol.~26, no.~12, pp. 5743--5757, 2017.

\bibitem{tii02}
D.~Li, L.~Deng, Z.~Cai, B.~Franks, and X.~Yao, ``Intelligent transportation
  system in macao based on deep self-coding learning,'' \emph{IEEE Trans Ind.
  Informat.}, vol.~14, no.~7, pp. 3253--3260, 2018.

\bibitem{tii03}
P.~Li, Z.~Chen, L.~T. Yang, Q.~Zhang, and M.~J. Deen, ``Deep convolutional
  computation model for feature learning on big data in internet of things,''
  \emph{IEEE Trans Ind. Informat.}, vol.~14, no.~2, pp. 790--798, 2018.

\bibitem{ai08}
Z.~Zhao, K.~M. Barijough, and A.~Gerstlauer, ``Deepthings: Distributed adaptive
  deep learning inference on resource-constrained iot edge clusters,''
  \emph{IEEE Trans. Computer-Aided Design Integr. Circuits Syst.}, vol.~37,
  no.~1, pp. 2348--2359, 2018.

\bibitem{ec01}
H.~Li, K.~Ota, and M.~Dong, ``Learning iot in edge: Deep learning for the
  internet of things with edge computing,'' \emph{IEEE Network}, vol.~32,
  no.~1, pp. 96--101, 2018.

\bibitem{tii04}
Z.~Zheng, Y.~Yang, X.~Niu, H.-N. Dai, and Y.~Zhou, ``Wide and deep
  convolutional neural networks for electricity-theft detection to secure smart
  grids,'' \emph{IEEE Trans Ind. Informat.}, vol.~14, no.~4, pp. 1606--1615,
  2018.

\bibitem{vs01}
H.~Kavalionak, C.~Gennaro, and G.~Amato, ``Distributed video surveillance using
  smart cameras,'' \emph{J. Grid Compu.}, pp. 1--19, 2018.

\bibitem{ai10}
H.~Han, A.~K. Jain, F.~Wang, S.~Shan, and X.~Chen, ``Heterogeneous face
  attribute estimation: A deep multi-task learning approach,'' \emph{IEEE
  Trans. Pattern Anal. Machine Intell.}, vol.~40, no.~11, pp. 2597--2609, 2018.

\bibitem{ai11}
D.~Cho, Y.-W. Tai, and I.~S. Kweon, ``Deep convolutional neural network for
  natural image matting using initial alpha mattes,'' \emph{IEEE Trans. Image
  Processing}, vol.~28, no.~3, pp. 1054--1067, 2019.

\bibitem{ai07}
J.~Chen, K.~Li, K.~Bilal, X.~Zhou, K.~Li, and P.~S. Yu, ``A bi-layered parallel
  training architecture for large-scale convolutional neural networks,''
  \emph{IEEE Trans. Parallel Distrib. Syst.}, pp. 1--1, 2018.

\bibitem{ai09}
M.~Langer, A.~Hall, Z.~He, and W.~Rahayu, ``Mpca sgd: A method for distributed
  training of deep learning models on spark,'' \emph{IEEE Trans. Parallel
  Distrib. Syst.}, vol.~29, no.~11, pp. 2540--2556, 2018.

\bibitem{vs02}
S.~Yi, Z.~Hao, Q.~Zhang, Q.~Zhang, W.~Shi, and Q.~Li, ``Lavea: latency-aware
  video analytics on edge computing platform,'' in \emph{Proceedings of the
  Second ACM/IEEE Symposium on Edge Computing}.\hskip 1em plus 0.5em minus
  0.4em\relax ACM, 2017, p.~15.

\bibitem{tii01}
L.~Li, K.~Ota, and M.~Dong, ``Deep learning for smart industry: Efficient
  manufacture inspection system with fog computing,'' \emph{IEEE Trans Ind.
  Informat.}, vol.~14, no.~10, pp. 4665--4673, 2018.

\bibitem{ec03}
A.~Diro and N.~Chilamkurti, ``Leveraging lstm networks for attack detection in
  fog-to-things communications,'' \emph{IEEE Commun. Mag.}, vol.~56, no.~9, pp.
  124--130, 2018.

\bibitem{ec02}
H.~Khelifi, S.~Luo, B.~Nour, A.~Sellami, H.~Moungla, S.~H. Ahmed, and
  M.~Guizani, ``Bringing deep learning at the edge of information-centric
  internet of things,'' \emph{IEEE Commun. Lett.}, pp. 1--1, 2018.

\bibitem{ai05}
D.~Zhang, W.~Wu, H.~Cheng, and R.~Zhang, ``Image-to-video person
  re-identification with temporally memorized similarity learning,'' \emph{IEEE
  Trans. Circuits Syst. Video Technol.}, vol.~28, no.~10, pp. 2622--2632, 2018.

\bibitem{tii06}
W.~Sun, J.~Liu, Y.~Yue, and H.~Zhang, ``Double auction-based resource
  allocation for mobile edge computing in industrial internet of things,''
  \emph{IEEE Trans Ind. Informat.}, vol.~14, no.~10, pp. 4692--4701, 2018.

\bibitem{vs03}
H.~D. Park and O.-G. Min, ``Scalable architecture for an automated surveillance
  system using edge computing,'' \emph{J. Supercomput}, vol.~73, no.~3, pp.
  926--939, 2017.

\bibitem{ec04}
S.~Yu, Y.~Wu, W.~Li, Z.~Song, and W.~Zeng, ``A model for fine-grained vehicle
  classification based on deep learning,'' \emph{Neurocomputing}, vol. 257, pp.
  97--103, 2017.

\bibitem{ec06}
X.~Ma, Z.~Tao, Y.~Wang, H.~Yua, and Y.~Wang, ``Long short-term memory neural
  network for traffic speed prediction using remote microwave sensor data,''
  \emph{Transportation Research}, vol.~54, pp. 187--197, 2015.

\bibitem{ec07}
Z.~Zhao, W.~Chen, X.~Wu, P.~C.~Y. Chen, and J.~Liu, ``Lstm network: a deep
  learning approach for short-term traffic forecast,'' \emph{IET Intelligent
  Transport Systems}, vol.~11, no.~2, pp. 68--75, 2017.

\end{thebibliography}

\begin{IEEEbiography}
[{\includegraphics[width=1in, height=1.25in, clip, keepaspectratio]{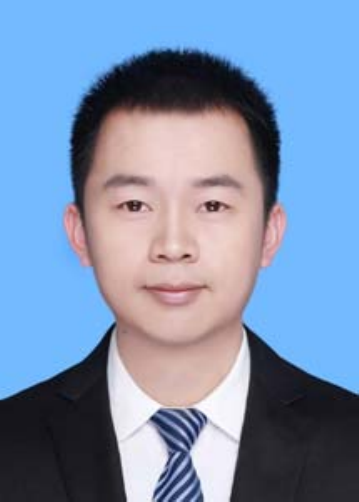}}]
{Jianguo Chen} received the Ph.D. degree in Computer Science and Technology from Hunan University, China, in 2018.
He was a visiting Ph.D. student at the University of Illinois at Chicago from 2017 to 2018.
He is currently a postdoctoral in University of Toronto and Hunan University.
His major research areas include parallel computing, cloud computing, machine learning, data mining, bioinformatics and big data.
He has published research articles in international conference and journals of data-mining algorithms and parallel computing, such as
{\em IEEE Transactions on Parallel and Distributed Systems},
{\em IEEE/ACM Transactions on Computational Biology and Bioinformatics},
{\em IEEE Transactions on Industrial Informatics},
and {\em Information Sciences}.
\end{IEEEbiography}

\begin{IEEEbiography}
[{\includegraphics[width=1in, height=1.25in, clip, keepaspectratio]{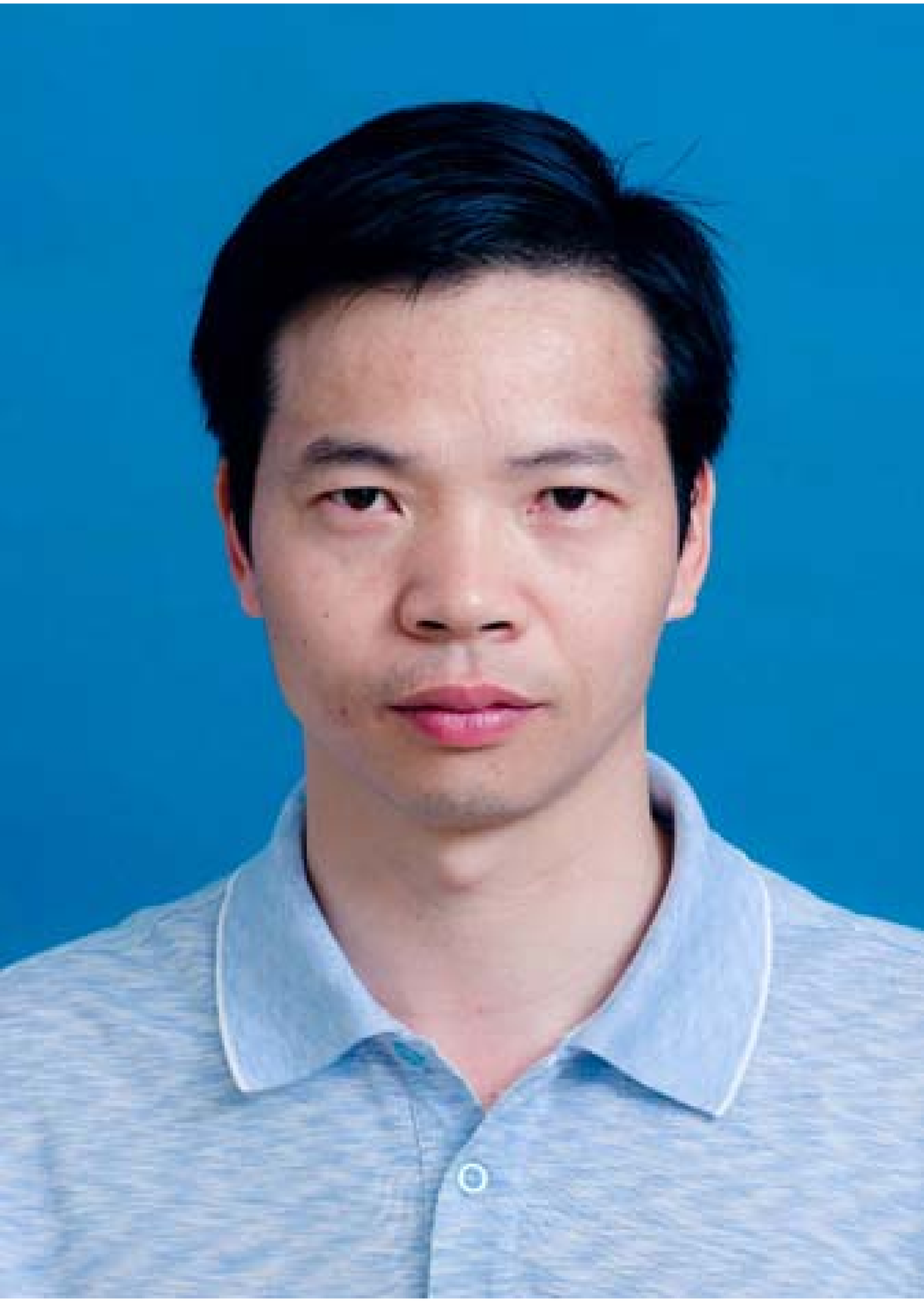}}]
{Kenli Li} received the Ph.D. degree in computer science from Huazhong University of Science and Technology, China, in 2003.
He was a visiting scholar at University of Illinois at Urbana-Champaign from 2004 to 2005.
He is currently a full professor of computer science and technology at Hunan University
and director of National Supercomputing Center in Changsha.
His major research areas include parallel computing, high-performance computing, grid and cloud computing.
He has published more than 200 research papers in international conferences and journals,
such as {\em IEEE-TC}, {\em IEEE-TPDS}, {\em IEEE-TSP}, {\em JPDC}, {\em ICPP}, {\em CCGrid}.
He is an outstanding member of CCF and a senior member of the IEEE, and serves on the editorial board of {\em IEEE Transactions on Computers}.
\end{IEEEbiography}

\begin{IEEEbiography}
[{\includegraphics[width=1in, height=1.25in, clip, keepaspectratio]{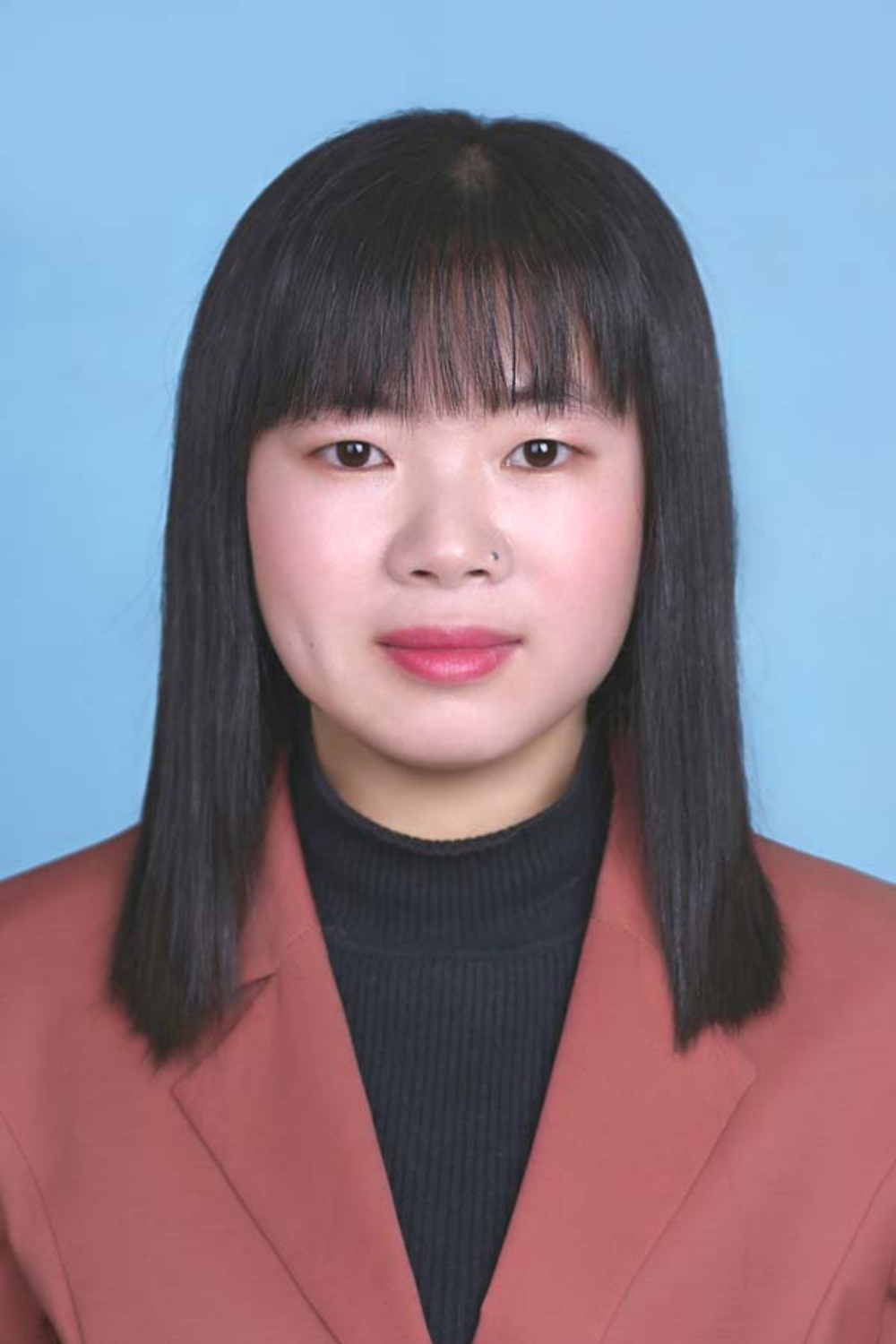}}]
{Qingying Deng} received the Ph.D. degree in Applied Mathematics from Xiamen University, China, in 2018.
She was a visiting Ph.D. student at the University of Illinois at Chicago from 2017 to 2018.
She is currently an assistant professor in the School of Mathematics and Computational Science at Xiangtan University, China.
Her major research areas include game theory, graph theory, graph mining, and machine learning.
She has published research articles in international conference and journals of mathematics and computer sciences, such as
{\em Linear Algebra and Its Application},
{\em Electronic Journal of Combinatorics},
{\em Discrete mathematics},
and {Journal of Knot Theory and Its Ramifications}.
\end{IEEEbiography}

\begin{IEEEbiography}
[{\includegraphics[width=1in, height=1.25in, clip, keepaspectratio]{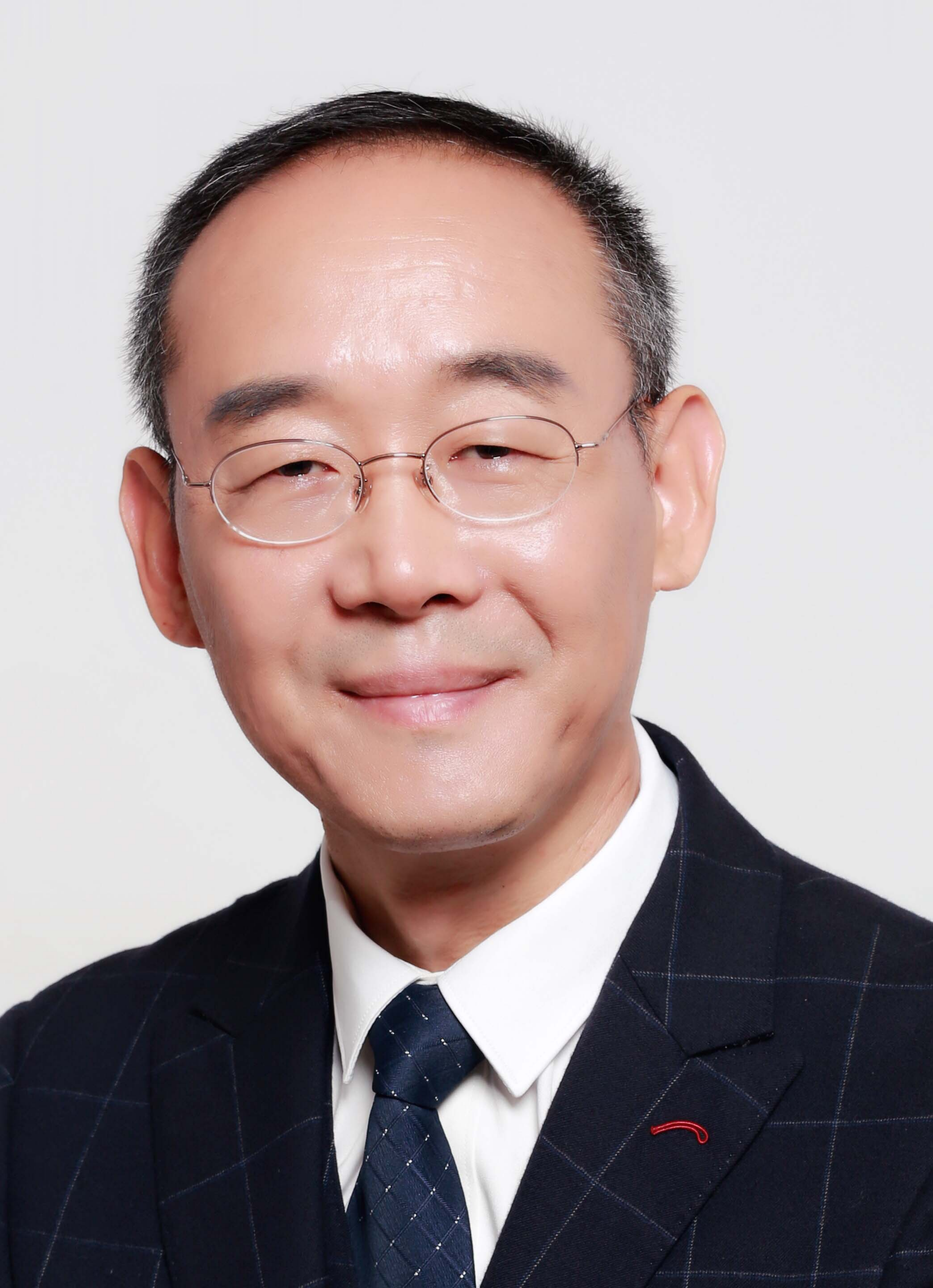}}]
{Keqin Li} is a SUNY Distinguished Professor of computer science with the State University of New York.
He is also a Distinguished Professor at Hunan University, China.
His current research interests include cloud computing, fog computing and mobile edge computing, energy-efficient computing and communication, embedded systems and cyberphysical systems, heterogeneous computing systems, big data computing, high-performance
computing, CPU-GPU hybrid and cooperative computing, computer architectures and systems, computer networking, machine learning, intelligent and soft computing.
He has published over 640 journal articles, book chapters, and refereed conference papers, and has received several best paper awards.
He currently serves or has served on the editorial boards of IEEE-TPDS, IEEE-TC, IEEE-TCC, and IEEE-TSC.
He is an IEEE Fellow.
\end{IEEEbiography}

\begin{IEEEbiography}
[{\includegraphics[width=1in, height=1.25in, clip, keepaspectratio]{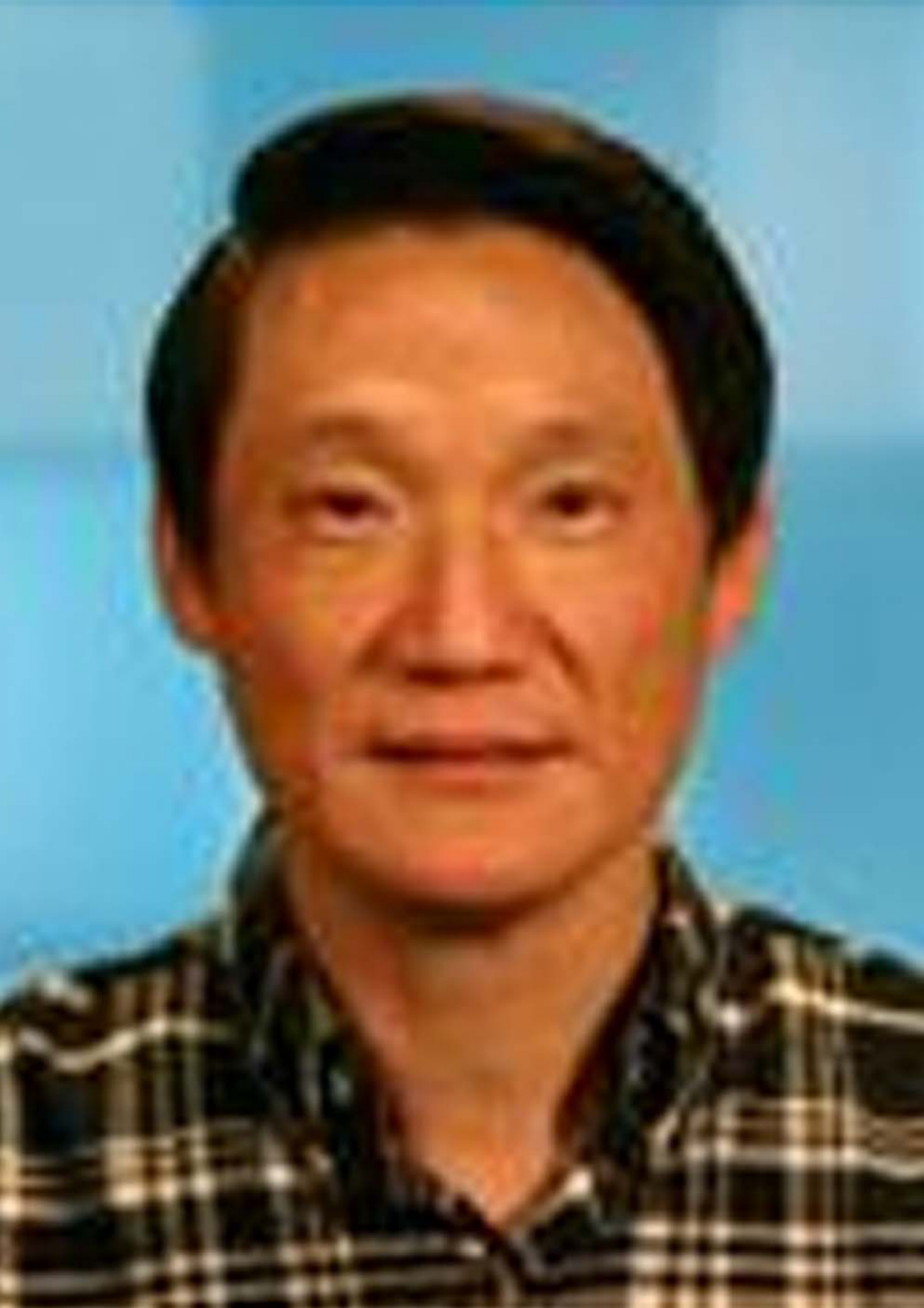}}]
{Philip S. Yu} received the B.S. Degree in E.E. from National Taiwan University, the M.S. and Ph.D. degrees in E.E. from Stanford University, and the M.B.A. degree from New York University.
He is a Distinguished Professor in Computer Science at the University of Illinois at Chicago and also holds the Wexler Chair in Information Technology.
His research interest is on big data, including data mining, data stream, database and privacy.
He has published more than 1,100 papers in refereed journals and conferences. He holds or has applied for more than 300 US patents.
He is the Editor-in-Chief of {\em ACM Transactions on Knowledge Discovery from Data}.
Dr. Yu is the recipient of ACM SIGKDD 2016 Innovation Award for his influential research and scientific contributions on mining, fusion and anonymization of big data, the IEEE Computer Society’s 2013 Technical Achievement Award for ``pioneering and fundamentally innovative contributions to the scalable indexing, querying, searching, mining and anonymization of big data'', and the Research Contributions Award from IEEE Intl. Conference on Data Mining (ICDM) in 2003 for his pioneering contributions to the field of data mining.
He also received the ICDM 2013 10-year Highest-Impact Paper Award, and the EDBT Test of Time Award (2014).
Dr. Yu is a Fellow of the ACM and the IEEE.
\end{IEEEbiography}
\end{document}